\documentclass[11pt,table,dvipsnames]{article}
\usepackage{tencent_tech_report}
\usepackage[colorlinks = true,
            linkcolor = blue,
            urlcolor  = blue,
            citecolor = blue,
            anchorcolor = blue]{hyperref}

\usepackage[most]{tcolorbox} 
\definecolor{tsinghuapurple}{RGB}{102,8,116}
\newtcolorbox{alprompt}[1]{
        boxrule = 1pt,
        fontupper = \small\tt,
        fonttitle = \bf\color{black},
        arc = 2pt,
        rounded corners,
        colframe = black,
        colbacktitle = white!97!yellow,
        colback = white!97!yellow,
        title = #1,
}

\usepackage{algorithmic}
\usepackage{algorithm}
\usepackage{amsmath}

\title{{\em Think Fast and Slow:} Step-Level Cognitive Depth Adaptation for LLM Agents}

\author{
Ruihan Yang$^\heartsuit$\thanks{Work done during an internship at Tencent Hunyuan.}, 
 Fanghua Ye$^\spadesuit$\thanks{Corresponding authors.}, 
 Xiang Wei$^\spadesuit$, 
 Ruoqing Zhao$^\spadesuit$, 
 Kang Luo$^\spadesuit$, 
 Xinbo Xu$^\heartsuit$,
 Bo Zhao$^\spadesuit$, \\ 
 \textbf{Ruotian Ma}$^\spadesuit$,\textbf{Shanyi Wang}$^\spadesuit$,
 \textbf{Zhaopeng Tu}$^\spadesuit$\footnotemark[2],
 \textbf{Xiaolong Li}$^\spadesuit$,
 \textbf{Deqing Yang}$^\heartsuit$\footnotemark[2], 
 \textbf{Linus}$^\spadesuit$\\
 $^\heartsuit$Fudan University \quad $^\spadesuit$Tencent Hunyuan \\
    \small \texttt{\{rhyang17\}@fudan.edu.cn} \quad
    \small \texttt{\{fanghua.ye.21\}@gmail.com}\\
    \quad \\
    Project Page: \url{https://github.com/rhyang2021/CogRouter}
}

\colmfinalcopy

\usepackage{microtype}
\usepackage{graphicx}
\usepackage{arydshln}
\usepackage{todonotes}
\usepackage{tabularx}
\usepackage{booktabs}
\usepackage{amsfonts}
\usepackage{multirow}
\usepackage{amssymb}
\usepackage{subcaption}

\usepackage[framemethod=TikZ]{mdframed} 
\usepackage{listings}
\usepackage{longtable}
\usepackage{xspace}
\usepackage{wrapfig}
\usepackage{pgfplots}
\usepackage{color, colortbl}
\usepackage{array}
\usepackage{paralist}
\usepackage{enumitem}
\usepackage[capitalize,noabbrev]{cleveref}

\newcommand{\rparagraph}[1]{\vspace{1.2mm}\noindent\textbf{#1.}}

\definecolor{Gray}{gray}{0.92}
\definecolor{racing-green}{rgb}{0.0, 0.8, 0.6}
\definecolor{awesome-red}{rgb}{1.0, 0.13, 0.32}
\definecolor{LightCyan}{rgb}{0.88,1,1}
\definecolor{darkgreen}{RGB}{0,150,0}
\definecolor{Ground}{RGB}{255,184,55}
\definecolor{Dirt}{RGB}{191,169,115}
\definecolor{Pink}{RGB}{226,184,176}
\definecolor{Violet}{RGB}{163,148,170}
\definecolor{darkred}{RGB}{150,0,0} %

\definecolor{lightblue}{RGB}{208,227,251}
\definecolor{level4}{RGB}{110,136,203}
\definecolor{level3}{RGB}{173,190,226}
\definecolor{level2}{RGB}{205,208,243}
\definecolor{level1}{RGB}{236,236,252}

\definecolor{lightgray}{gray}{0.95}
\definecolor{lightblue}{RGB}{230,240,255}
\newcolumntype{g}{>{\columncolor{Ground!7}}c}
\newcolumntype{d}{>{\columncolor{cyan!6}}c}
\newcolumntype{f}{>{\columncolor{lime!6}}c}
\newcolumntype{v}{>{\columncolor{purple!6}}c}

\newcommand{\ie}{\textit{i}.\textit{e}.,\ }
\newcommand{\eg}{\textit{e}.\textit{g}.,\ }
\newcommand{\framework}{\textsc{CogRouter}\xspace}
\newcommand{\cosft}{\textsc{CoSFT}\xspace}
\newcommand{\copo}{\textsc{CoPO}\xspace}

\begin{document}
\maketitle
\begin{abstract}
Large language models (LLMs) are increasingly deployed as autonomous agents for multi-turn decision-making tasks. However, current agents typically rely on fixed cognitive patterns: non-thinking models generate immediate responses, while thinking models engage in deep reasoning uniformly. This rigidity is inefficient for long-horizon tasks, where cognitive demands vary significantly from step to step, with some requiring strategic planning and others only routine execution. In this paper, we introduce \textbf{\textsc{CogRouter}}, a framework that trains agents to dynamically adapt cognitive depth at each step. Grounded in ACT-R theory, we design four hierarchical cognitive levels ranging from instinctive responses to strategic planning. Our two-stage training approach includes \textbf{Cognition-aware Supervised Fine-tuning} (\cosft) to instill stable level-specific patterns, and \textbf{Cognition-aware Policy Optimization} (\copo) for step-level credit assignment via confidence-aware advantage reweighting. 
The key insight is that appropriate cognitive depth should maximize the confidence of the resulting action. Experiments on ALFWorld and ScienceWorld demonstrate that \framework achieves state-of-the-art performance with superior efficiency. 
With Qwen2.5-7B, it reaches an 82.3\% success rate, outperforming GPT-4o (+40.3\%), OpenAI-o3 (+18.3\%), and GRPO (+14.0\%), while using 62\% fewer tokens.
\end{abstract}

\section{Introduction}

Recent advances in large language models (LLMs)~\citep{gpt4, Gemini2023} have enabled their deployment as autonomous agents capable of planning and executing complex multi-turn tasks across domains, including code generation~\citep{yuan2024easytoolenhancingllmbasedagents,lu2025codetool}, software engineering~\citep{jimenez2024swebench}, web interaction~\citep{yao2023webshop, zhou2024webarenarealisticwebenvironment}, and embodied environments~\citep{yang2025lighthouselanguageenhancingllm}. 
In these settings, agents must continuously perceive observations, reason about actions, and execute decisions over long horizons. Unlike single-turn tasks where a fixed thinking strategy may suffice, agentic environments are highly dynamic and impose {\bf varying cognitive demands} at each step within a task.

Consider the ``use thermometer'' task in ScienceWorld~\citep{wang2022scienceworlda}, where the agent must first locate the thermometer and then measure the temperature of an unknown substance. In the initial exploration phase, the agent must plan strategically, deciding which rooms to search and in what order to efficiently locate the required objects. 
When encountering unexpected situations, such as searching the workshop but finding no thermometer, the agent needs to reflect on the failed search and revise its exploration plan. In contrast, many mid-trajectory steps are routine. After executing ``open door to bathroom'' and observing that the door opens, the natural next action is simply ``go to bathroom'' without extended thinking. This illustrates that agentic tasks exhibit significant {\bf step-wise heterogeneity} in thinking demands, ranging from instinctive reactions to strategic planning.

\begin{wrapfigure}{r}{0.5\textwidth}
\centering
\includegraphics[width=0.48\textwidth]{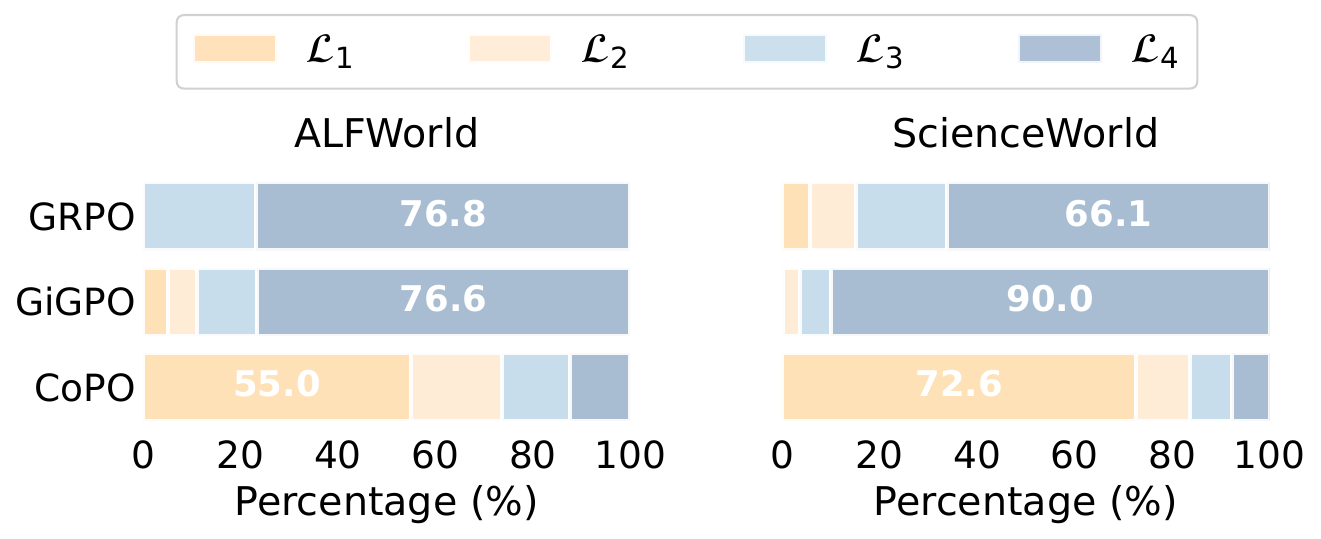}
\caption{Illustration of the {\bf cognitive rigidity issue}: While \copo maintains an adaptive cognitive distribution (bottom), standard RL methods like GRPO (top) collapse to uniform deep thinking ($\mathcal{L}_4$), wasting computational resources on routine steps. $\mathcal{L}_1$–$\mathcal{L}_4$ represent increasing cognitive depth, from instinctive responses to strategic planning.}
\label{fig:rigidity}
\end{wrapfigure}

However, most agents rely on {\bf fixed cognitive patterns}. Non-thinking models produce reflexive actions at every step, while thinking models (\eg DeepSeek-R1~\citep{deepseekai2025deepseekr1incentivizingreasoningcapability}) engage in deep chain-of-thought reasoning throughout the entire trajectory. This rigidity creates a critical inefficiency: thinking models consume excessive tokens on routine steps where simple responses suffice, while non-thinking models fail on complex decisions requiring strategic planning. 
A natural solution is to train agents that adaptively allocate cognitive resources. 
However, as shown in Figure~\ref{fig:rigidity}, even when initialized with balanced cognitive capabilities, standard reinforcement learning (RL) methods tend to collapse toward uniform deep thinking. 
While recent work has explored adaptive thinking for single-turn tasks~\citep{lou2025adacotparetooptimaladaptivechainofthought,zhang2025adaptthinkreasoningmodelslearn}, these methods typically employ task-level adaptation with binary switching, failing to capture the {\bf step-wise, multi-level} cognitive 
demands within agentic tasks.

This raises a central challenge: {\em How can we train agents to adaptively allocate cognitive depth across multiple levels at each step, balancing both task performance and computational efficiency?} To address this, we introduce \textsc{CogRouter}, a framework grounded in the Adaptive Control of Thought-Rational (ACT-R) theory~\citep{anderson1982acquisition}, which posits that humans dynamically allocate cognitive resources from automatic procedural execution to deliberate reasoning. \textsc{CogRouter} trains agents to modulate their cognitive depth across {\bf four hierarchical levels}: from instinctive responses ($\mathcal{L}_1$) to strategic planning ($\mathcal{L}_4$).

Training such adaptive allocation is non-trivial due to the sparsity of rewards in agentic environments. We propose a two-stage approach: {\em Cognition-aware Supervised Fine-tuning} (\cosft) to instill stable, level-specific cognitive patterns, followed by {\em Cognition-aware Policy Optimization} (\copo). \copo is a novel RL algorithm that enables step-level credit assignment via confidence-aware advantage reweighting. The core insight is that an appropriate cognitive depth should facilitate confident action prediction.

We evaluate \framework on two challenging interactive benchmarks, ALFWorld~\citep{shridhar2020alfworld} and ScienceWorld~\citep{wang2022scienceworlda}. Experimental results demonstrate that \framework achieves state-of-the-art performance with superior token efficiency. Using Qwen2.5-7B as the base model, our approach achieves an 82.3\% average success rate, significantly outperforming GPT-4o (+40.3\%), OpenAI-o3 (+18.3\%), and GRPO (+14.0\%), while reducing token consumption by 62\% compared to GRPO.

In summary, our contributions are three-fold:
\begin{enumerate}[leftmargin=10pt]
    \item We identify and formalize the ``cognitive rigidity'' issue in LLM agents, illustrating how fixed reasoning patterns lead to inefficiency in long-horizon tasks with heterogeneous step-wise demands.
    \item We propose \framework, a framework grounded in ACT-R theory that defines four hierarchical cognitive levels and utilizes a two-stage training pipeline (\cosft and \copo) to enable dynamic cognitive adaptation.
    \item We introduce \copo, a novel RL algorithm that performs step-level credit assignment via confidence-aware advantage reweighting, allowing the model to learn efficient reasoning without collapsing to uniform deep thinking.
\end{enumerate}

\section{Preliminary}
\paragraph{Partially Observable Markov Decision Process} 
In this paper, we model the agentic task as a partially observable Markov decision process (POMDP) $\mathcal{M} = (\mathcal{X}, \mathcal{S}, \mathcal{A}, \mathcal{O}, \mathcal{T}, \mathcal{R})$, where $\mathcal{X}$ denotes the set of instructions, $\mathcal{S}$ represents the set of environment states, $\mathcal{A}$ is the action space, and $\mathcal{O}$ is the observation space. 
The state transition function is $\mathcal{T}: \mathcal{S} \times \mathcal{A} \rightarrow \Delta(\mathcal{S})$, and the reward function is $\mathcal{R}: \mathcal{S} \times \mathcal{A} \rightarrow \mathbb{R}$.
At each time step $t$, the agent with a policy $\pi_\theta$ receives an observation $o_t \in \mathcal{O}$, which is correlated with the current state $s_t \in \mathcal{S}$. 
The agent maintains an interaction history $\tau_t = \{x, (o_0, a_0), \ldots, (o_{t-1}, a_{t-1}), o_t\}$ and samples an action $a_t \sim \pi_\theta(\cdot \mid \tau_t)$, which induces a state transition $s_{t+1} \sim \mathcal{T}(\cdot \mid \tau_t, a_t)$. 
An episode terminates at step $T$, yielding trajectory
$\tau = \{x, (o_0, a_0), \ldots, (o_T, a_T)\}$ and terminal reward $\mathcal{R}(\tau)$.
The agent’s objective is to learn a policy $\pi_\theta$ that maximizes the expected terminal reward:
$$
\theta^* = \arg\max_{\theta} \mathbb{E}_{\substack{a_t \sim \pi_\theta(\cdot \mid \tau_t)}} [\mathcal{R}(\tau)].
$$

\paragraph{GRPO for Agentic Tasks} 
Group Relative Policy Optimization (GRPO)~\citep{shao2024deepseekmathpushinglimitsmathematical} is an effective method for training models without the need for a separate critic network. 
In agentic tasks, GRPO operates at the trajectory level. Given a task query $x \in \mathcal{X}$, the agent samples a set of $G$ complete trajectories $\{\tau^{(i)}\}_{i=1}^G$ using the policy $\pi_{\theta}$. 
Each trajectory $\tau^{(i)} = \{x, (o_0^{(i)}, a_0^{(i)}), \dots, (o_T^{(i)}, a_T^{(i)})\}$ begins with the instruction $x$, followed by a sequence of observation-action pairs over $T$ time steps.
The trajectory-level reward $R_i = \mathcal{R}(\tau^{(i)})$ measures overall task success. 
The advantage for each trajectory is computed as
\begin{equation}
    \label{eq:grpo_adv}
    A^{(i)} = \frac{R_i - \text{mean}\left(\{R_1, R_2, \dots, R_G\}\right)}{\text{std}\left(\{R_1, R_2, \dots, R_G\}\right)}.
\end{equation}
This trajectory-level advantage $A^{(i)}$ is uniformly assigned to all steps within the trajectory, serving as the learning signal for policy optimization. The objective function for GRPO is
\begin{equation*}
\begin{aligned}    \mathcal{J}_{\text{GRPO}}(\theta) 
    = &\mathbb{E}_{x \sim \mathcal{X}, \{\tau^{(i)}\}_{i=1}^G \sim \pi_{\theta_{\text{old}}}(\cdot|x)} \\
    &\left[\frac{1}{G} \sum_{i=1}^G \frac{1}{|a^{(i)}|} \sum_{t=0}^{T} \sum_{n=1}^{|a_t^{(i)}|} \min\left(r_{t,n}^{(i)}(\theta) A^{(i)}, \text{clip}(r_{t,n}^{(i)}(\theta), 1-\epsilon, 1+\epsilon) A^{(i)}\right)\right] - \beta \mathbb{D}_{\mathrm{KL}}\left[\pi_\theta \| \pi_{\mathrm{ref}}\right],
\end{aligned}
\end{equation*}
where $|a^{(i)}| = \sum_{t=0}^{T} |a_t^{(i)}|$ is the total number of action tokens in trajectory $\tau^{(i)}$, and $r_{t,n}^{(i)}(\theta) = \frac{\pi_\theta(a_{t,n}^{(i)} \mid x, \tau_{t}^{(i)}, a_{t,<n}^{(i)})}{\pi_{\theta_{\text{old}}}(a_{t,n}^{(i)} \mid x, \tau_{t}^{(i)}, a_{t,<n}^{(i)})}$ denotes the importance sampling ratio for the $n$-th token in the action $a_t^{(i)}$.
\section{Methodology}  

\begin{figure*}
    \centering    \includegraphics[width=0.999\linewidth]{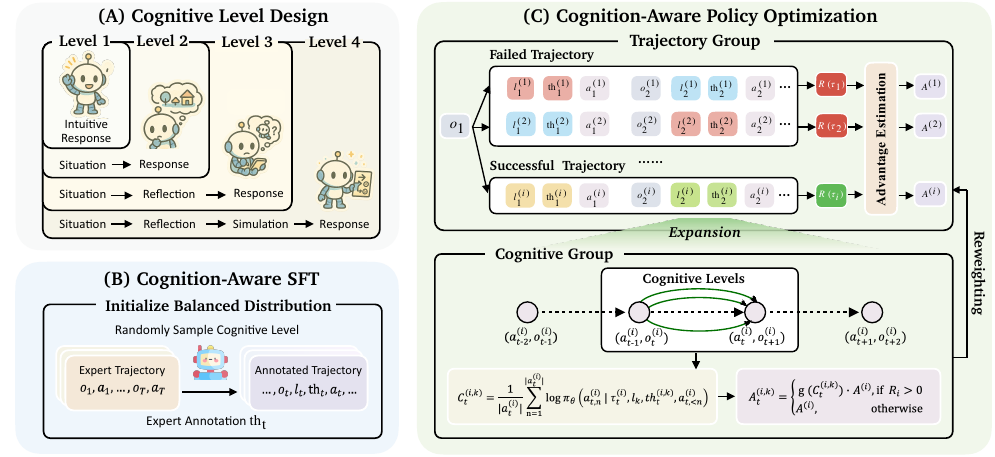}
    \caption{Overview of the \framework framework. We define four cognitive levels $\mathcal{L}_{1}$–$\mathcal{L}_{4}$, then introduce a two-stage training process: (1) \textbf{Cognition-aware Supervised Fine-tuning (\cosft)}, which guides the model to learn stable cognitive patterns across levels with balanced data; (2) \textbf{Cognition-aware Policy Optimization (\copo)}, which applies RL with confidence-aware reweighting to help the model adaptively choose suitable levels based on context complexity.
}
    \label{fig:main}
\end{figure*}

We propose \framework (Figure~\ref{fig:main}), a framework for training models to dynamically select the appropriate cognitive level at each step. 
We first define a set of distinct \textbf{cognitive levels}, grounded in ACT-R theory (\S~\ref{sec:level_design}).
Building on this design, we introduce a two-stage training pipeline:
\begin{inparaenum}[\it 1)]
\item \textbf{Cognition-aware Supervised Fine-tuning} (\cosft), which integrates level-specific cognitive patterns into the model's base capabilities (\S\ref{sec:cogsft}); and
\item \textbf{Cognition-aware Policy Optimization} (\copo), 
which optimizes cognitive level selection through reinforcement 
learning with confidence-aware advantage reweighting (\S\ref{sec:copo}).
\end{inparaenum}

\subsection{Task Formulation}
\label{sec:task_formulation}
In a standard Partially Observable Markov Decision Process (POMDP), the agent generates actions directly from the interaction history. However, complex agentic tasks often demand deeper reasoning at specific steps. 
To address this, we introduce $N$ \textbf{cognitive levels}, denoted as $\mathcal{L} = \{\mathcal{L}_1, \dots, \mathcal{L}_N\}$, where each level $\mathcal{L}_i$ represents a distinct depth of reasoning. At each time step $t$, the agent with policy $\pi_\theta$, first selects a cognitive level $l_t \in \mathcal{L}$ based on the interaction history: $l_t \sim \pi_\theta(\cdot \mid \tau_t)$. Given $l_t$, it generates an intermediate thinking process: $th_t \sim \pi_\theta(\cdot \mid \tau_t, l_t)$, and then produces an executable action conditioned on this reasoning: $a_t \sim \pi_\theta(\cdot \mid \tau_t, th_t)$.
We denote the complete structured output at step $t$ as $y_t = [l_t, th_t, a_t]$. The objective is to maximize the expected terminal reward:
\[
\theta^* = \arg\max_{\theta} \mathbb{E}_{l_t \sim \pi_\theta(\cdot \mid \tau_t), \, th_t \sim \pi_\theta(\cdot \mid \tau_t, l_t), \, a_t \sim \pi_\theta(\cdot \mid \tau_t, th_t)} \left[\mathcal{R}(\tau)\right].
\]
This formulation requires the agent to adaptively select an appropriate cognitive level based on the complexity of the current state and to generate corresponding reasoning, balancing depth with efficiency.\footnote{Note that the reward $\mathcal{R}(\tau)$ depends jointly on the cognitive level $l_t$, reasoning process $th_t$, and action $a_t$, and does not necessarily increase with reasoning depth.}

\subsection{Cognitive Level Design}
\label{sec:level_design}
Adaptive Control of Thought-Rational (\textbf{ACT-R})~\citep{anderson1982acquisition, anderson1993rules} posits that human cognition operates across a spectrum from automatic procedural execution to effortful declarative reasoning, with processing depth inversely related to cognitive load. 
Motivated by this framework, we propose \textbf{four cognitive levels} for agent decision-making that dynamically balance cognitive depth with computational efficiency (see full prompt in Appendix~\ref{app:cogrouter_prompt}).

\noindent\textbf{Level 1 $\mathcal{L}_1$ (Instinctive Response)}: $\mathcal{L}_1$ represents autonomous processing, corresponding to ACT-R's procedural stage where compiled production rules execute automatically without working memory engagement~\citep{anderson1998atomic}. 
The agent produces \textit{immediate responses} based on learned patterns, with no explicit reasoning structure. This mode optimizes for speed in routine scenarios.

\noindent\textbf{Level 2 $\mathcal{L}_2$ (Situational Awareness)}: $\mathcal{L}_2$ introduces monitored execution with basic deliberation. 
The agent assesses the \textit{Current State} and \textit{Available Actions} before selecting a response. 
This level aligns with ACT-R's goal-directed procedural processing, where situational information is maintained in working memory to guide action selection~\citep{taatgen2013nature}.

\noindent\textbf{Level 3 $\mathcal{L}_3$ (Experience Integration)}: $\mathcal{L}_3$ engages retrospective reasoning by incorporating historical information. Beyond situational assessment, the agent specifies the \textit{Goal}, performs \textit{Reflection} on past actions and outcomes, and integrates experiential insights into decision-making. This corresponds to ACT-R's knowledge compilation stage, where declarative memories are retrieved and consolidated to refine procedural knowledge~\citep{anderson1982acquisition}.

\noindent\textbf{Level 4 $\mathcal{L}_4$ (Strategic Planning)}: $\mathcal{L}_4$ represents the most cognitively demanding mode, requiring extensive \textit{prospective simulation}. 
The agent evaluates multiple candidate actions by mentally simulating their future consequences and impact. 
This level mirrors ACT-R's fully declarative stage, engaging complex problem-solving through chunk retrieval and strategic evaluation~\citep{anderson2007human}.

\subsection{Cognition-Aware Supervised Fine-tuning}
\label{sec:cogsft}
While prior work on controllable reasoning often emphasizes the reinforcement learning (RL) stage, we argue that the supervised fine-tuning (SFT) stage is equally critical for establishing a solid foundation:
\begin{inparaenum}[\it 1)]
\item The model should first acquire stable cognitive patterns for each cognitive level to prevent format leakage or mode collapse during subsequent RL training; and
\item The model may inherently favor certain cognitive patterns, which can interfere with learning to select levels based on task complexity. Enforcing a balanced distribution over all cognitive levels during SFT helps mitigate this bias.
\end{inparaenum}

To implement the structured output $y_t = [l_t, th_t, a_t]$ 
defined in \S~\ref{sec:task_formulation}, we introduce a structured format. Each step $t$ includes three components: 
\begin{inparaenum}[\it 1)]
\item the cognitive level, enclosed in \texttt{<level>} tags;
\item the internal thinking process corresponding to the selected level, enclosed in \texttt{<think>} tags;
\item the final executable action, enclosed in \texttt{<action>} tags.
\end{inparaenum}
Using this structured format, the training data is constructed 
by augmenting expert trajectories with cognitive annotations 
as follows:
\begin{enumerate}[leftmargin=*]
    \item We collect a dataset of successful trajectories $\mathcal{T}^{*} = \{(x, o_0, a^{*}_0, \ldots, o_T, a^{*}_T)\}$, comprising only observation-action pairs generated by expert model (\ie GPT-4o).
    \item At each step $t$, we randomly sample a cognitive level $l_t$ to ensure balanced coverage across all levels.
    \item Given the history $\tau^{*}_t = \{x, (o_0, a^{*}_0), \dots, (o_{t-1}, a^{*}_{t-1}), o_t\}$ and the ground-truth action $a^{*}_t$, we prompt the expert model to complete the thinking process $th_t$ corresponding to the selected level $l_t$ (see Appendix~\ref{app:cogrouter_prompt} for the detailed prompt).
\end{enumerate}

This procedure yields a balanced dataset $\mathcal{D}_{\rm{cog}} = \{(\tau^*_t, l_t, \text{th}_t, a^*_t)\}_{t=0}^{T}$ with approximately uniform distribution across four cognitive levels. 
The SFT objective minimizes the negative log-likelihood over all steps:
\begin{equation*}
\mathcal{L}_{\mathrm{CoSFT}} = -\mathbb{E}_{\mathcal{D}_{\rm{cog}}}\left[\sum_{t=0}^{T} \log \pi_\theta\left(y_t \mid \tau^*_t\right)\right],
\end{equation*}
where $\pi_\theta$ is the policy parameterized by $\theta$, and $y_t = [l_t, \text{th}_t, a^*_t]$ denotes the structured output. 
As an ablation, we also construct an alternative dataset where the expert model directly selects the cognitive level at each step rather than random sampling.

\begin{algorithm}[h]
\caption{Cognition-Aware Policy Optimization (\copo)}
\label{algo:copo}
\begin{algorithmic}[1]
\STATE \textbf{Initialize:} Policy $\pi_\theta$.
\FOR{each training iteration}
    \STATE Collect a batch of trajectories $\mathcal{B} = \{\tau^{(i)}\}$ by running policy $\pi_\theta$.
    \STATE Compute trajectory-level rewards $R_i$ and advantages $A^{(i)}$ for each $\tau^{(i)} \in \mathcal{B}$.
    \FOR{each successful trajectory $\tau^{(i)}$ with $R_i > 0$}
        \FOR{each step $t$ in trajectory $\tau^{(i)}$}
            \STATE Construct cognitive group $e_t^{(i)}$ by generating thinking processes under all 4 levels.
            \STATE Compute action prediction confidence $C_t^{(i,k)}$ for each level $k$ using Eq.~\ref{eq:confidence}.
            \STATE Normalize confidence scores $C_{\text{norm},t}^{(i,k)}$ within the cognitive group using Eq.~\ref{eq:norm_confidence}.
            \STATE Compute confidence-aware weights $g(C_t^{(i,k)})$ using Eq.~\ref{eq:weight}.
            \STATE Compute step-level advantages $A_t^{(i,k)}$ using Eq.~\ref{eq:step_adv}.
        \ENDFOR
    \ENDFOR
    \STATE For failed trajectories ($R_i \leq 0$), keep original advantages without cognitive group expansion.
    \STATE Update policy $\pi_\theta$ by maximizing the \copo objective (Eq.~\ref{eq:copo_objective}).
\ENDFOR
\STATE \textbf{Output:} Optimized policy $\pi_\theta$.
\end{algorithmic}
\end{algorithm}

\subsection{Cognition-Aware Policy Optimization}
\label{sec:copo}
While group-based RL algorithms (\eg GRPO) have proven highly effective for training LLMs on reasoning tasks, they assign trajectory-level advantages uniformly across all steps, without distinguishing whether the cognitive pattern at each step is contextually appropriate. 
To address this, we propose \textbf{Cognition-Aware Policy Optimization (\copo)}, which enables step-level credit assignment based on action prediction confidence (Figure~\ref{fig:main}). The key insight is that \textit{an appropriate cognitive pattern should facilitate confident and correct action selection.}
\copo modulates step-level advantages using the model's prediction confidence, allowing it to learn which cognitive depth is appropriate for each step. The algorithm is shown in Algorithm~\ref{algo:copo}.

\paragraph{Reward Design}
Given trajectory $\tau^{(i)}$, the trajectory-level reward $R_i$ consists of two components:
\begin{equation*}
R_i = R^{\text{task}}_i \times R^{\text{format}}_i,
\end{equation*}
where $R^{\text{task}}_i \in \{0, 1\}$ indicates task success based on environment feedback, 
and $R^{\text{format}}_i \in \{0, 1\}$ indicates whether all actions in the trajectory conform 
to the required structured format (\ie, properly using \texttt{<level>}, \texttt{<think>}, 
and \texttt{<action>} tags). 
We set $R^{\text{format}}_i = 0$ if any action deviates from 
this format, imposing a strict penalty to enforce format consistency.

\paragraph{Cognitive Group Construction} 
For each successful trajectory ($R_i > 0$), we expand the training data by constructing a \textbf{cognitive group} at each step. Specifically, we regenerate the thinking process under all four cognitive levels while keeping the observation $o_t^{(i)}$ and action $a_t^{(i)}$ fixed. 
This yields the cognitive group $e_t^{(i)} = \{e_t^{(i,k)}\}_{k=1}^4$, where each variant is defined as $e_t^{(i,k)} = [l_k, th_t^{(i,k)}, a_t^{(i)}]$, corresponding to cognitive level $l_k$.
Importantly, all variants produce the same action $a_t^{(i)}$, but differ in the thinking process that leads to it. To assess the suitability of each cognitive level, we evaluate how confidently the model predicts the action when conditioned on each thinking process:
\begin{equation}
\label{eq:confidence}
C_t^{(i,k)} = \frac{1}{|a_t^{(i)}|} \sum_{n=1}^{|a_t^{(i)}|} 
\log \pi_\theta\left(a_{t,n}^{(i)} \mid \tau_t^{(i)}, l_k, th_t^{(i,k)}, a_{t,<n}^{(i)}\right),
\end{equation}
where $a_{t,n}^{(i)}$ is the $n$-th token in the action $a_t^{(i)}$.
Higher confidence scores indicate stronger alignment between the thinking process and the resulting action.

\paragraph{Confidence-Aware Advantage Reweighting}
To compare cognitive levels within each group, we normalize the confidence scores:
\begin{equation}
\label{eq:norm_confidence}
C_{\text{norm},t}^{(i,k)} = \frac{C_t^{(i,k)} - \mu_t^{(i)}}{\sigma_t^{(i)}},
\end{equation}
where $\mu_t^{(i)}$ and $\sigma_t^{(i)}$ are the mean and standard deviation of confidence scores within the cognitive group $e_t^{(i)}$.
We then apply a temperature-scaled softmax to convert these normalized scores into relative weights
\begin{equation}
\label{eq:weight}
g(C_t^{(i, k)}) = \frac{\exp(m \cdot C_{\text{norm},t}^{(i,k)})}{\sum_{j=1}^4 \exp(m \cdot C_{\text{norm},t}^{(i,j)})},
\end{equation}
where $m$ is a temperature parameter that controls the sharpness of the distribution\footnote{We set $m=2$ in our experiments.}. 
Since $\sum_{k=1}^4 g(C_t^{(i,k)}) = 1$ by construction, these weights redistribute the total advantage across cognitive levels without changing its magnitude.
These weights are used to modulate the step-level advantage as follows:
\begin{equation}
\label{eq:step_adv}
A_t^{(i, k)} = \begin{cases}
g(C_t^{(i, k)}) \cdot A^{(i)}, & \text{if } R_i > 0 \\
A^{(i)}, & \text{otherwise}
\end{cases}
\end{equation}
where $A^{(i)}$ is the trajectory-level advantage (Eq.~\ref{eq:grpo_adv}).
For successful trajectories ($R_i > 0$), this reweighting amplifies advantages for cognitive levels that facilitate confident action predictions, while attenuating uncertain ones.
For failed trajectories, no cognitive group is constructed, and the advantage remains at the trajectory level.

\paragraph{\copo Optimization}
For successful trajectories ($i \in \mathcal{I}^+$), we use the 
cognitive groups $e_t^{(i)}$ with reweighted advantages $A_t^{(i,k)}$ for each level from Equation~\ref{eq:step_adv}.
For failed trajectories ($i \in \mathcal{I}^-$), we retain only the original structured output $y_t^{(i)}$ with trajectory-level advantage $A^{(i)}$.
The objective function is
\begin{equation}
\label{eq:copo_objective}
\begin{aligned}
\mathcal{J}_{\text{CoPO}}(\theta) = &\mathbb{E}_{x \sim \mathcal{X}, \{\tau^{(i)}\}_{i=1}^G \sim \pi_{\theta_{\text{old}}}(\cdot|x)} \Bigg[
\frac{1}{G} \Bigg( \sum_{i \in \mathcal{I}^+} \frac{1}{|e^{(i)}|} \sum_{t=0}^{T} \sum_{k=1}^{4} \sum_{n=1}^{|e_t^{(i,k)}|} \min\Big(r_{t,n}^{(i,k)} \hat{A}_t^{(i,k)}, \overline{r}_{t,n}^{(i,k)} \hat{A}_t^{(i,k)}\Big) \\
&+ \sum_{i \in \mathcal{I}^-} \frac{1}{|y^{(i)}|} \sum_{t=0}^{T} \sum_{n=1}^{|y_t^{(i)}|}  \min\Big(r_{t,n}^{(i)} \hat{A}^{(i)}, \overline{r}_{t,n}^{(i)} \hat{A}^{(i)}\Big) \Bigg) \Bigg] - \beta \mathbb{D}_{\text{KL}}[\pi_\theta \| \pi_{\text{ref}}],
\end{aligned}
\end{equation}
where $|e^{(i)}| = \sum_{t,k} |e_t^{(i,k)}|$ and $|y^{(i)}| = \sum_{t} |y_t^{(i)}|$ are the total token counts for normalization, $r_{t,n}^{(i,k)} = \frac{\pi_\theta(e_{t,n}^{(i,k)} \mid x, \tau_{t}^{(i)}, e_{t,<n}^{(i,k)})}{\pi_{\theta_{\text{old}}}(e_{t,n}^{(i,k)} \mid x, \tau_{t}^{(i)}, e_{t,<n}^{(i,k)})}$ and 
$r_{t,n}^{(i)} = \frac{\pi_\theta(y_{t,n}^{(i)} \mid x, \tau_{t}^{(i)}, y_{t,<n}^{(i)})}{\pi_{\theta_{\text{old}}}(y_{t,n}^{(i)} \mid x, \tau_{t}^{(i)}, y_{t,<n}^{(i)})}$ 
denote the importance sampling ratios for successful and failed trajectories respectively, 
and $\overline{r} = \text{clip}(r, 1{-}\epsilon, 1{+}\epsilon)$ is the clipped ratio.

\section{Experiments}
\subsection{Experimental Setup}

\paragraph{Benchmarks}
We conduct experiments on two challenging environments:
\begin{inparaenum}[\it 1)]
\item \textbf{ALFWorld}~\citep{shridhar2020alfworld} is an embodied environment designed to evaluate LLM agents on multi-step decision-making across six categories of household tasks: Pick \& Place (Pick), Examine in Light (Look), Clean \& Place (Clean), Heat \& Place (Heat), Cool \& Place (Cool), and Pick Two \& Place (Pick2). 
In each task, the agent receives a textual goal and must complete it through multi-turn interaction with the environment. 
A task is considered successful if the goal is achieved within the maximum number of allowed steps.
\item \textbf{ScienceWorld}~\citep{wang2022scienceworlda} is a text-based environment that tests scientific reasoning across 30 task types aligned with the elementary science curriculum. At the end of each task, the agent receives a score between 0 and 100 based on performance.
\end{inparaenum}
Further details can be found in Appendix~\ref{app:eval_settings}.

\paragraph{Evaluation}
Following prior work~\citep{zhang2025rlvmr,xi2024agentgymevolvinglargelanguage}, we evaluate models on the test sets of both environments: 200 simulations for ALFWorld and the same for ScienceWorld. 
For ALFWorld, we report the success rate for each task category and overall. For ScienceWorld, we use both the average final score and success rate as the evaluation metrics.
We also compute the average output token usage per task (\textbf{\#Tokens}) as a measure of model efficiency.

\paragraph{Training Settings}
We use Llama3.1-8B~\citep{llama3modelcard} and Qwen2.5-7B~\citep{yang2024qwen2technicalreport} as base models, with training data drawn from ScienceWorld (2,120 simulations) and ALFWorld (2,420 simulations). 
For \cosft, we randomly sample 500 environments from each benchmark, collect expert trajectories using GPT-4o~\citep{gpt-4o}, and prompt it to complete thinking processes at randomly sampled cognitive levels (\S~\ref{sec:cogsft}).
For \copo, we use the remaining simulations, sample different 16 groups per rollout with a group size of 8, and run 150 iterations.
Additional details are provided in Appendix~\ref{app:implementation}.

\paragraph{Baselines}
We compare our method against several baselines (see Appendix~\ref{app:baseline} for implementation details):
\begin{itemize}[leftmargin=12pt]
\item \textbf{Prompt-based methods}, including ReAct~\citep{yao2023react} and Reflexion~\citep{shinn2023reflexion}, a self-refinement method that summarizes each trial to guide future decisions. We also evaluate R1-distilled variants of Llama3.1-8B and Qwen2.5-7B, prompted with ReAct;
\item \textbf{Training-based methods}, covering both offline and online methods. Offline methods include SFT and ETO~\citep{song2024trialerrorexplorationbasedtrajectory}. We also include cognition-aware SFT (\cosft; see \S\ref{sec:cogsft}), which uniformly samples cognitive levels during training, and its variant (\textsc{CoSFT}$_{\text{exp}}$), which uses expert-selected levels based on contextual complexity.
Online methods include GRPO~\citep{shao2024deepseekmathpushinglimitsmathematical},  GiGPO~\citep{feng2025groupingrouppolicyoptimizationllm}, and AdaptThink~\citep{zhang2025adaptthinkreasoningmodelslearn}.
\end{itemize}
We also include several non-thinking and thinking frontier models for comparison.

\begin{table*}[t!]
\centering
\small
\caption{Performance comparison across different methods on ALFWorld and ScienceWorld. \textbf{SR}: Success Rate (\%). \textbf{\#Tokens}: Average tokens per trajectory. \copo achieves state-of-the-art performance with superior token efficiency across both benchmarks.}
\setlength{\tabcolsep}{6.7pt}
\scalebox{0.93}{
\begin{tabular}{l rr rrr rr}
\toprule
\multirow{2}{*}{\textbf{Method}} & \multicolumn{2}{c}{\textbf{ALFWorld}} & \multicolumn{3}{c}{\textbf{ScienceWorld}} & \multicolumn{2}{c}{\textbf{Average}}\\
\cmidrule(lr){2-3} \cmidrule(lr){4-6} \cmidrule(lr){7-8}
& \textbf{SR} & \textbf{\#Tokens} & \textbf{Score} & \textbf{SR} & \textbf{\#Tokens} & \textbf{SR} & \textbf{\#Tokens}\\
\midrule
\rowcolor{lightgray}
\multicolumn{8}{c}{\textbf{Frontier Large Models}} \\
\addlinespace[0.25em]
\texttt{gpt-4o-2024-08-06}~\citep{gpt-4o} & 61.5 & 860.8 & 57.0 & 22.5 & 1009.9 & 42.0 & 935.4\\
\texttt{deepseek-v3-0324}~\citep{deepseekai2025deepseekv3technicalreport} & 52.0 & 1171.2 & 13.0 & 6.0 & 1013.4 & 29.0 & 1092.3\\
\addlinespace[0.2em]
\cdashline{1-8}
\addlinespace[0.3em]
\texttt{openai-o3}~\citep{openai2025o3mini} & 74.0 & 4290.9 & 72.4 & 54.0 & 5184.0 & 64.0 & 4737.5\\
\texttt{deepseek-r1-0528}~\citep{deepseekai2025deepseekr1incentivizingreasoningcapability} & 73.0 & 6230.9 & 58.9 & 40.0 & 16502.7 & 56.5 & 11366.8\\
\texttt{gemini-2.5-pro-0506}~\citep{comanici2025gemini25pushingfrontier} & 70.0 & 4896.5 & 65.2 & 44.0 & 8942.4 & 57.0 & 6919.5 \\
\midrule
\rowcolor{lightgray}
\multicolumn{8}{c}{\textbf{Qwen2.5-7B-Instruct}} \\
\addlinespace[0.25em]
ReAct~\citep{yao2023react} & 50.5 & 666.4 & 8.0 & 2.5 & 1645.1 & 26.5 & 1155.8\\
~~~+ R1-Distill & 4.5 & 33620.5 & 2.5 & 0.5 & 21221.9 & 2.5 & 27421.2\\
Reflexion~\citep{shinn2023reflexion} & 42.5 & 2340.8 & 10.1 & 1.0 & 3396.8 & 21.8 & 2868.8\\
\addlinespace[0.1em]
\hline
\addlinespace[0.25em]
SFT & 62.5 & 494.1 & 54.1 & 19.0 & 660.8 & 40.8 & 577.5\\
ETO~\citep{song2024trialerrorexplorationbasedtrajectory} & 69.5 & 540.6 & 53.9 & 21.5 & 694.3 & 45.5 & 617.4\\
AdaptThink~\citep{zhang2025adaptthinkreasoningmodelslearn} & 81.5 & 1372.1 & 62.5 & 44.5 & 1314.2 & 63.0 & 1343.2\\
\textsc{CoSFT}$_{\text{exp}}$ & 78.5 & 3151.0 & 61.2 & 29.5 & 3973.8 & 54.0 & 3562.4\\
\addlinespace[0.2em]
\cdashline{1-8}
\addlinespace[0.3em]
\cosft & 57.0 & 3912.7 & 61.1 & 35.5 & 3643.4 & 46.3 & 3778.1\\
~~~+ GRPO~\citep{shao2024deepseekmathpushinglimitsmathematical} & 83.5 & 4994.6 & 71.1 & 53.0 & 3739.9 & 68.3 & 4367.3\\
~~~+ GiGPO~\citep{feng2025groupingrouppolicyoptimizationllm} & 88.0 & 2955.5 & 67.3 & 47.0 & 4602.8 & 67.5 & 3779.2\\
\addlinespace[0.1em]
\rowcolor{lightblue}
~~~\textbf{+ \copo (Ours)} & \textbf{92.5} & 1739.4 & \textbf{84.6} & \textbf{72.0} & 1543.4 & \textbf{82.3} & 1641.4\\
\midrule
\rowcolor{lightgray}
\multicolumn{8}{c}{\textbf{Llama3.1-8B-Instruct}} \\
\addlinespace[0.25em]
ReAct~\citep{yao2023react} & 13.0 & 959.4 & 13.7 & 3.0 & 1268.1 & 8.0 & 1113.8\\
~~~+ R1-Distill & 2.5 & 21720.8 & 2.2 & 0.0 & 6940.8 & 1.3 & 14330.8\\
Reflexion~\citep{shinn2023reflexion} & 18.5 & 3645.8 & 17.8 & 2.5 & 5072.4 & 10.5 & 4359.1\\
\addlinespace[0.1em]
\hline
\addlinespace[0.25em]
SFT & 54.5 & 632.0 & 59.9 & 24.0 & 906.6 & 39.3 & 769.3\\
ETO~\citep{song2024trialerrorexplorationbasedtrajectory} & 65.0 & 730.0 & 62.5 & 38.5 & 784.2 & 51.8 & 757.1\\
AdaptThink~\citep{zhang2025adaptthinkreasoningmodelslearn} & 79.5 & 1607.2 & 52.3 & 45.5 & 1666.8 & 62.5 & 1637.0\\
\textsc{CoSFT}$_{\text{exp}}$ & 82.0 & 2903.2 & 73.3 & 49.5 & 4588.9 & 65.8 & 3746.1\\
\addlinespace[0.2em]
\cdashline{1-8}
\addlinespace[0.3em]
\cosft & 55.5 & 3546.6 & 56.4 & 26.0 & 3728.0 & 40.8 & 3637.3\\
~~~+ GRPO~\citep{shao2024deepseekmathpushinglimitsmathematical} & 84.0 & 3941.1 & 75.1 & 50.5 & 5757.6 & 67.3 & 4849.4\\
~~~+ GiGPO~\citep{feng2025groupingrouppolicyoptimizationllm} & 89.5 & 6112.0 & 78.2 & 64.0 & 4651.3 & 76.8 & 5381.7\\
\addlinespace[0.1em]
\rowcolor{lightblue}
~~~\textbf{+ \copo (Ours)} & \textbf{91.5} & 811.6 & \textbf{83.7} & \textbf{70.5} & 963.8 & \textbf{81.0} & 887.7\\
\bottomrule
\end{tabular}
}
\label{tab:main_results}
\end{table*}

\subsection{Main Results}
\label{sec:main_results}
We compare \framework against state-of-the-art frontier models and both offline and online training methods. 
Key findings are summarized below.

\rparagraph{Fixed cognitive patterns and binary switching are insufficient for long-horizon agentic tasks}
As shown in Table~\ref{tab:main_results}, models with fixed cognitive patterns struggle on agentic tasks.
Non-thinking models like DeepSeek-V3 achieve only 29.0\% average success rate across both benchmarks (1092.3 tokens), while thinking models like DeepSeek-R1 reach 56.5\% but consume nearly 10$\times$ more tokens (11366.8).
The problem is exacerbated in smaller models: vanilla Qwen2.5-7B achieves merely 26.5\% (1155.8 tokens), while its R1-Distill variant collapses to 2.5\% despite using 24$\times$ more tokens (27421.2).
AdaptThink~\citep{zhang2025adaptthinkreasoningmodelslearn}, which employs binary switching between thinking and non-thinking modes, substantially improves over fixed-pattern baselines. However, it still underperforms \textsc{CogRouter} by 19.3\% on Qwen2.5-7B and 18.5\% on LLama3.1-8B.
This performance gap arises from two core limitations:
\begin{inparaenum}[\it 1)]
\item binary switching cannot capture multi-level cognitive demands, some steps require intermediate reasoning such as situational awareness or experience integration;
\item designed for single-turn tasks, AdaptThink uses a 
trajectory-level reward that uniformly encourages non-thinking 
across all steps, pushing the model toward selecting non-thinking 
regardless of step-wise complexity.
\end{inparaenum}
These findings underscore the need for multi-level adaptive cognition and step-wise credit assignment to effectively handle long-horizon agentic tasks.

\begin{figure*}[t]
    \centering
       \vspace{-0.1cm}    \includegraphics[width=1\linewidth]{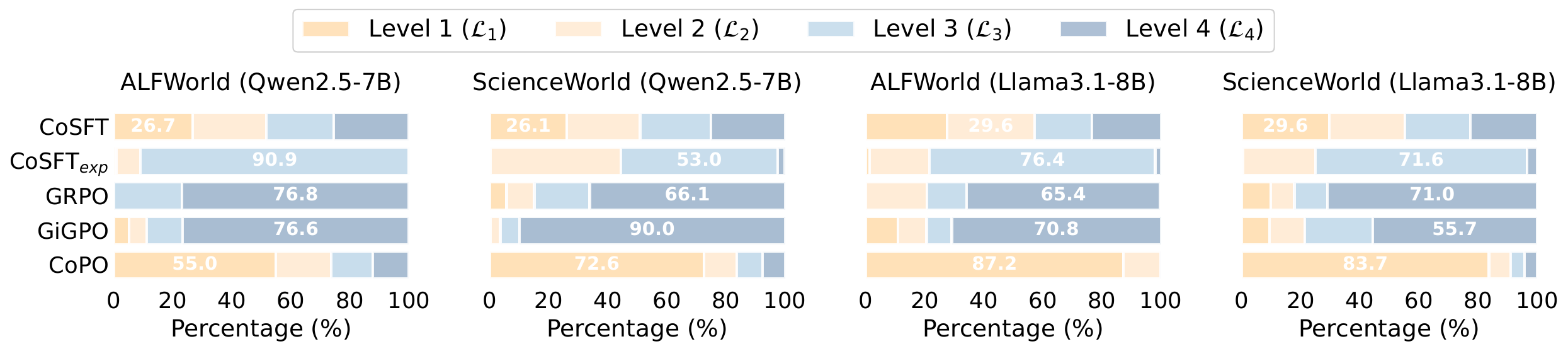}
    \caption{Cognitive level distribution after training. All RL methods (GRPO, GiGPO, \copo) are initialized from \cosft. While GRPO and GiGPO collapse to predominantly $\mathcal{L}_4$ thinking, \copo learns adaptive allocation.}
    \label{fig:main_01}
    \vspace{-0.4cm}
\end{figure*}

\rparagraph{Supervised fine-tuning alone cannot learn adaptive thinking}
While \textsc{CoSFT}$_{\text{exp}}$ achieves stronger performance than SFT (+13.2\% average success rate with Qwen2.5-7B), it consumes significantly more tokens (3562.4 compared to 577.5).
As shown in Figure~\ref{fig:main_01}, \textsc{CoSFT}$_{\text{exp}}$ inherits strong biases toward $\mathcal{L}_2$ and $\mathcal{L}_3$ thinking from the expert model. This highlights a core weakness of supervised fine-tuning: it mimics the expert's thinking distribution without learning to adapt cognitive depth based on context. 
Reinforcement learning is therefore essential for discovering step-wise optimal cognitive patterns.

\begin{figure*}[t]
    \centering   \includegraphics[width=1\linewidth]{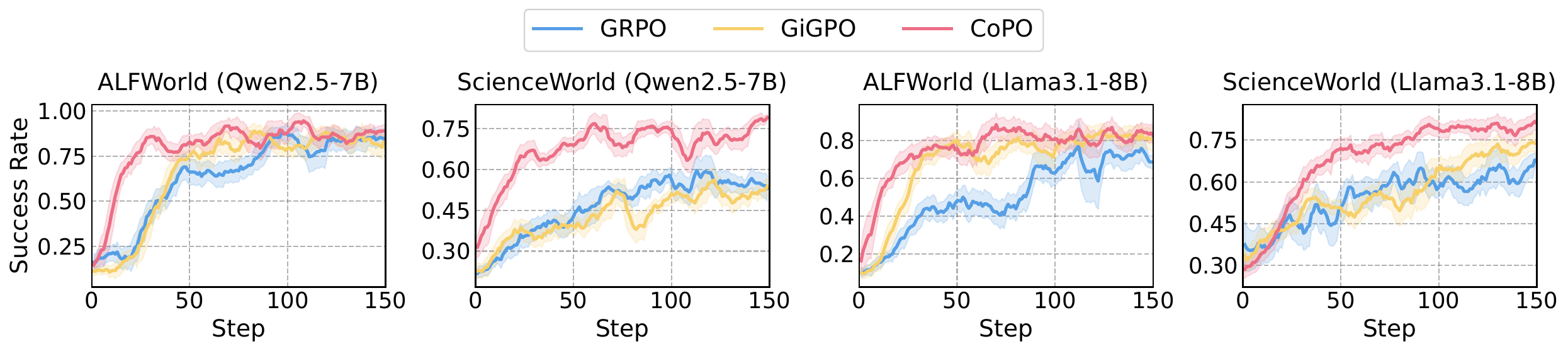}
    \caption{Training curves showing success rate across RL iterations for GRPO, GiGPO and \copo on ALFWorld and ScienceWorld. \copo achieves faster convergence to higher success rates.}
    \label{fig:main_02}
\end{figure*}

\rparagraph{CoPO prevents format collapse through confidence-aware credit assignment}
RL baselines struggle to adapt cognitive depth at the step level (Figure~\ref{fig:main_01}). 
Although all methods begin with the same \cosft initialization and a balanced level distribution, GRPO and GiGPO quickly collapse to predominantly $\mathcal{L}_4$ thinking (\eg 76.8\% and 76.6\% $\mathcal{L}_4$ on ALFWorld with Qwen2.5-7B). 
This collapse stems from coarse credit assignment: GRPO distributes advantages uniformly across all tokens, failing to distinguish whether deep thinking is contextually appropriate at each step.
GiGPO's state-based grouping also struggles, as identical states may require different cognitive depths depending on task progression. 
As a result, both methods converge to fixed $\mathcal{L}_4$ thinking, since deeper thinking often correlates with higher final rewards.
CoPO prevents this collapse via confidence-aware advantage reweighting (\S\ref{sec:copo}), which evaluates each cognitive level based on how confidently the model predicts its action under that cognitive pattern. 
This enables step-wise adaptive thinking, applying deeper thinking only when beneficial (\eg 55.0\% $\mathcal{L}_1$, 18.7\% $\mathcal{L}_2$, 14.0\% $\mathcal{L}_3$, 12.3\% $\mathcal{L}_4$ on ALFWorld with Qwen2.5-7B).
Moreover, CoPO converges substantially faster than prior RL methods (Figure~\ref{fig:main_02}). 
With Qwen2.5-7B, it reaches 90\% success rate on ALFWorld within 100 steps, whereas GRPO plateaus at 83.3\% after 150 steps; on ScienceWorld, CoPO reaches 76.6\% by step 80 compared to GRPO's 59.7\% ceiling. 
Similar trends appear for Llama3.1-8B.
This acceleration arises from CoPO's step-level, confidence-guided credit assignment, which provides more precise learning signals than trajectory-level baselines. 

\rparagraph{\framework achieves state-of-the-art performance with superior token efficiency}
Table~\ref{tab:main_results} shows that \framework consistently outperforms all baselines across both benchmarks. 
With Qwen2.5-7B, it achieves an average success rate of 82.3\%, surpassing larger frontier models such as GPT-4o (+40.3\%) and OpenAI-o3 (+18.3\%), as well as both offline and online methods, including ETO (+36.8\%) and GRPO (+14.0\%). 
Similar gains are observed with Llama3.1-8B, where \framework reaches 81.0\%, outperforming GRPO by 13.7\% and GiGPO by 4.2\%. 
Notably, \framework achieves this with superior token efficiency: using 62\% fewer tokens than GRPO (1641.4 compared to 4367.3) and 57\% fewer than GiGPO (1641.4 compared to 3779.2) with Qwen2.5-7B. 
These results highlight that adaptive cognitive depth is both more effective and more efficient than fixed cognitive patterns in long-horizon agentic tasks.

\section{Analysis} \subsection{Quantitative Analysis} While \S~\ref{sec:main_results} demonstrates \copo's superior performance and efficiency, its underlying mechanism remains unclear. To reveal why \copo succeeds, we analyze the learned cognitive level distributions along two dimensions: 
\begin{inparaenum}[\it 1)] 
\item \textbf{Trajectory-level}: how cognitive patterns evolve across trajectory stages, and 
\item \textbf{Task-level}: how cognitive depth scales with task complexity.
\end{inparaenum} 

\begin{figure*}[t] 
\centering 
\vspace{-0.1cm} \includegraphics[width=1\linewidth]{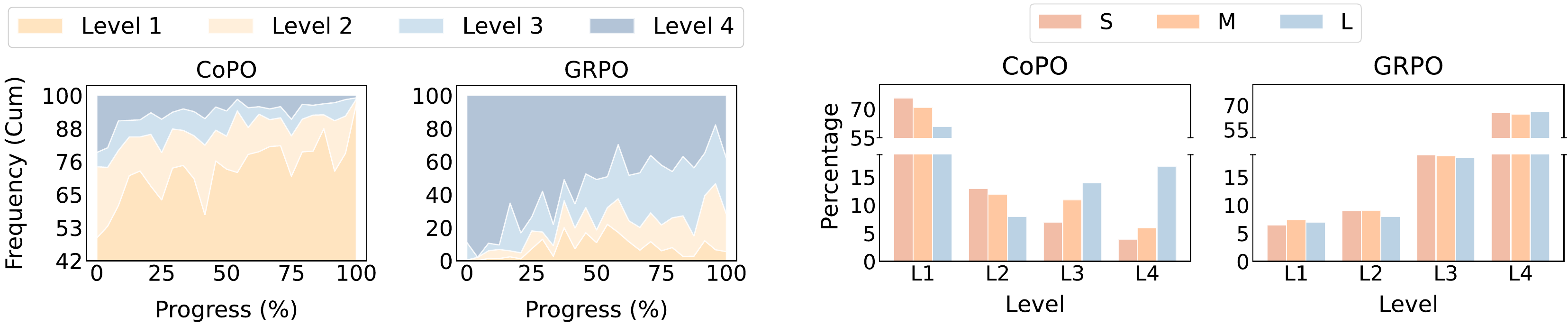} \caption{Cognitive level distributions across trajectory progress (left) and task complexity (right) for \copo and GRPO on ScienceWorld. 
} 
\label{fig:main_04} 
\vspace{-0.1cm} 
\end{figure*}

\paragraph{\copo learns to allocate cognitive resources dynamically based on the trajectory stage.}
We examine how the four cognitive levels are distributed across task progress in ScienceWorld, using Qwen2.5-7B as the base model (Figure~\ref{fig:main_04}, left).
\copo exhibits distinct, contextually appropriate patterns for each level. $\mathcal{L}_4$ peaks at initialization (21.9\% at 0\% progress) when complex tasks require global goal evaluation and long-horizon planning, then declines to 6.8\% as task structure becomes clear. 
$\mathcal{L}_2$ also dominates initial stages (26.3\% at 0\% progress), where environmental awareness is critical for parsing observations and assessing available actions, then stabilizes around 10\% as the agent maintains situational monitoring. 
As trajectories progress, $\mathcal{L}_1$ surges from 48.5\% initially to over 80\% in late stages, reaching 94.8\% at task completion with minimal $\mathcal{L}_3$-$\mathcal{L}_4$ usage, reflecting that most steps become routine executions requiring no deliberation. 
$\mathcal{L}_3$ maintains relatively uniform distribution around 10\% across all stages, emerging contextually only when errors occur or past experience must inform decisions. 
This structured allocation is cognitively plausible: strategic planning dominates initialization, situational awareness guides exploration, routine execution increases as procedures crystallize, and reflection appears when needed. 
In contrast, GRPO saturates trajectories with $\mathcal{L}_4$ thinking from start to finish (over 50\% throughout), collapsing into uniform deep reasoning regardless of step complexity.\footnote{GiGPO similarly collapses to $\mathcal{L}_4$ thinking. See Appendix~\ref{app:cognitive_collapse} for detailed analysis.}
This leads to higher token consumption without performance gains.
Detailed case studies are provided in Appendix~\ref{app:case_study}.

\paragraph{\copo adapts cognitive depth proportionally to task complexity.}
To investigate how cognitive levels adapt across tasks of varying complexity, we compare \copo and GRPO on ScienceWorld with Qwen2.5-7B as base model (Figure~\ref{fig:main_04}, right). 
We categorize tasks into Short (S), Medium (M), and Long (L) based on oracle trajectory lengths, where longer trajectories indicate higher complexity (see Appendix~\ref{app:eval_settings} for details). 
\copo demonstrates clear complexity-aware adaptation: as task difficulty increases from S to L, $\mathcal{L}_4$ rises from 7.6\% to 22.4\% and $\mathcal{L}_3$ increases from 8.6\% to 15.2\%, reflecting greater need for strategic planning and experiential reflection in complex scenarios. 
In contrast, $\mathcal{L}_1$ decreases from 72.6\% to 55.4\% and $\mathcal{L}_2$ drops from 11.2\% to 7.0\%, as instinctive responses can handle fewer steps when tasks become complex. 
This graduated allocation reveals that \copo scales cognitive effort proportionally with task complexity. In contrast, GRPO maintains nearly uniform distributions across all difficulty levels. The percentages of $\mathcal{L}_1$, $\mathcal{L}_2$, $\mathcal{L}_3$, and $\mathcal{L}_4$ remain stable from short to long tasks, indicating an inability to adapt cognitive allocation to actual task demands.

\subsection{Ablation Study}
\label{sec:ablation}

To validate the key design choices of \framework, we conduct ablation studies as follows: 
\begin{inparaenum}[\it 1)]
\item the confidence metric for advantage reweighting in \copo, and
\item training strategies for cognitive level distribution.
\end{inparaenum}
Results are presented in Table~\ref{tab:ablation_confidence}.

\definecolor{lightgray}{gray}{0.95}
\definecolor{lightblue}{RGB}{230,240,255}
\begin{table*}[t]
\centering
\small
\caption{Ablation studies on confidence metrics and training strategies. SR: Success Rate (\%). \#Tokens: Average tokens per trajectory. $\mathcal{L}_1$-$\mathcal{L}_4$: Cognitive level distribution (\%).}
\label{tab:ablation_confidence}
\setlength{\tabcolsep}{5.2pt}
\scalebox{0.98}{
\begin{tabular}{lcccccccccccc}
\toprule
\multirow{2}{*}{\textbf{Variant}} & \multicolumn{6}{c}{\textbf{ALFWorld}} & \multicolumn{6}{c}{\textbf{ScienceWorld}} \\
\cmidrule(lr){2-7} \cmidrule(lr){8-13}
& \textbf{SR} & \textbf{\#Tokens} & \textbf{$\mathcal{L}_1$} & \textbf{$\mathcal{L}_2$} & \textbf{$\mathcal{L}_3$} & \textbf{$\mathcal{L}_4$} & \textbf{SR} & \textbf{\#Tokens} & \textbf{$\mathcal{L}_1$} & \textbf{$\mathcal{L}_2$} & \textbf{$\mathcal{L}_3$} & \textbf{$\mathcal{L}_4$} \\
\midrule
\rowcolor{lightgray}
\multicolumn{13}{c}{\textbf{Confidence Metric Ablation}} \\
\addlinespace[0.25em]
\copo (Ours) & \textbf{92.5} & 1739.4 & 55.0 & 18.7 & 14.0 & 12.3 & \textbf{72.0} & \textbf{1543.4} & 72.6 & 11.2 & 8.6 & 7.6 \\
\ \ w/ \emph{Max Probs} & 89.0 & 1664.9 & 65.7 & 12.3 & 11.4 & 8.4 & 69.5 & 2167.5 & 68.3 & 15.2 & 9.5 & 7.1 \\
\ \ w/ \emph{Min Probs} & 81.5 & 1227.7 & 92.7 & 0.7 & 2.6 & 4.0 & 67.5 & 2063.6 & 69.5 & 12.7 & 10.7 & 7.1 \\
\ \ w/ \emph{Entropy} & 79.0 & 1584.6 & 73.9 & 11.5 & 6.3 & 8.3 & 70.0 & 1873.9 & 72.0 & 12.7 & 8.6 & 6.7 \\
\midrule
\rowcolor{lightgray}
\multicolumn{13}{c}{\textbf{Training Strategy Ablation}} \\
\addlinespace[0.25em]
\ \ w/o  \emph{Cold Start} & 86.0 & 2496.2 & 0.3 & 35.0 & 33.5 & 31.2 & 64.5 & 3750.6 & 0.0 & 27.1 & 35.5 & 37.3 \\
\ \ w/ \emph{\textsc{CoSFT}$_{\rm{expert}}$} & 87.5 & 2871.6 & 0.0 & 7.9 & 92.1 & 0.0 & 62.0 & 4383.0 & 0.0 & 24.3 & 69.1 & 0.6 \\
\ \ w/ \emph{Corr \& Incorr} & 83.0 & 1375.8 & 83.1 & 4.3 & 7.7 & 4.8 & 66.0 & 1755.6 & 77.6 & 9.9 & 6.9 & 5.6 \\
\bottomrule
\end{tabular}
}
\end{table*}

\paragraph{Average log-probability is the most effective confidence metric for assessing cognitive suitability.}
We evaluate our choice of average log-probability (\S\ref{sec:copo}) as the confidence metric by comparing three alternatives: \emph{Max Probs} uses the maximum log-probability across action tokens, \emph{Min Probs} uses the minimum, and \emph{Entropy} measures distributional uncertainty over the vocabulary. 
All variants are trained with Qwen2.5-7B initialized from \cosft.
As shown in Table~\ref{tab:ablation_confidence}, average log-probability yields the strongest performance (92.5\% SR on ALFWorld, 72.0\% on ScienceWorld).
\emph{Min Probs} performs substantially worse (81.5\%, 67.5\%), as a single low-probability token dominates the confidence score regardless of overall prediction quality, creating unreliable credit assignment. 
\emph{Entropy} similarly degrades performance (79.0\%, 70.0\%), as it measures distributional uncertainty over the full vocabulary rather than whether the cognitive pattern supports the action being executed.
\emph{Max Probs} achieves moderate results (89.0\%, 69.5\%) but overweights local peaks, ignoring the global coherence of the action sequence.
These results validate average log-probability as the optimal confidence metric for COPO.

\paragraph{Balanced initialization and success-only reweighting are crucial for effective adaptation.}
\label{sec:ablation_training}
We examine two critical design decisions in \copo training: using balanced \cosft for cold-start initialization, and applying confidence-based reweighting only to successful trajectories. 
We compare against three variants (Table~\ref{tab:ablation_confidence}).
Both \textit{w/o Cold Start} (directly applying RL without SFT) and \textit{w/ CoSFT$_{\text{expert}}$} (initializing from GPT-4o's expert-selected levels) demonstrate the necessity of balanced distribution. 
Without cold start, the model exhibits severe distribution collapse with extreme skew (\eg 0.3\% $\mathcal{L}_1$ on ALFWorld, 0.0\% on ScienceWorld). 
With expert initialization, it inherits teacher preferences, collapsing to $\mathcal{L}_3$ (92.1\%) while ignoring $\mathcal{L}_1$ and $\mathcal{L}_4$ (0.0\%). Both prevent effective exploration during RL, as the model either fails to establish stable cognitive formatting or becomes trapped in biased patterns. 
This leads to degraded performance (86.0\% and 87.5\% compared to 92.5\%) and increased token usage (2496.2 and 2871.6 compared to 1739.4).
Applying reweighting to both successful and failed trajectories (\textit{w/ Corr \& Incorr}) creates adverse incentives. Failed trajectories receive stronger penalties for high confidence, encouraging the model to avoid certainty. 
The model exploits this by collapsing to $\mathcal{L}_1$ (83.1\%) to evade "confident mistakes", since $\mathcal{L}_1$'s minimal thinking yields lower confidence and lighter penalties. 
This reduces tokens (1375.8) but sacrifices performance (83.0\%).
In contrast, \copo achieves 92.5\% SR with 1739.4 tokens and adaptive distribution.

\section{Related Work}
\rparagraph{Interactive and Agentic Environments}
Recent studies have introduced evaluation tasks for language agents requiring 
long-horizon planning and strategic reasoning in multi-turn, goal-driven
settings, including code generation~\citep{jimenez2024swebench, kimiteam2026kimik25visualagentic}, 
embodied intelligence~\citep{xi2024agentgymevolvinglargelanguage, shridhar2020alfworld, wang2022scienceworlda}, 
social deduction games~\citep{yang2024selfgoallanguageagentsknow, yang2025ariatraininglanguageagents, huang2025fardecisionmakingllmsevaluating}, 
and emotional intelligence~\citep{zhang2025sentientagentjudgeevaluating, chen2024personapersonalizationsurveyroleplaying, yi2025goodbadfailurellms}. 
Unlike single-turn reasoning tasks, these environments feature sparse 
terminal rewards over long horizons and heterogeneous step-wise 
complexity.
Some decisions require strategic planning while many are routine executions. This motivates adaptive cognitive depth allocation rather than uniform cognitive patterns.

\rparagraph{Hybrid and Adaptive Reasoning}
Recent large reasoning models such as OpenAI-o1~\citep{openai2025o3mini}, 
DeepSeek-R1~\citep{deepseekai2025deepseekr1incentivizingreasoningcapability}, and Gemini-2.0-Flash-Thinking~\citep{comanici2025gemini25pushingfrontier} demonstrate impressive capabilities through extended chain-of-thought reasoning. However, they apply uniform deep thinking throughout trajectories. To address efficiency, recent work explores adaptive thinking~\citep{zhang2025adaptthinkreasoningmodelslearn,lou2025adacotparetooptimaladaptivechainofthought,yang2025qwen3technicalreport}, 
enabling models to dynamically switch between thinking and non-thinking modes based on problem difficulty. 
While effective for single-turn reasoning tasks, 
these binary approaches cannot capture the step-wise, multi-level cognitive demands within agentic tasks, where some steps require situational awareness or experience integration beyond simple binary switching. 
Moreover, they operate at the trajectory level, uniformly incentivizing efficiency across all steps rather than adapting to step-specific complexity. 
We propose step-level adaptive cognitive allocation grounded in ACT-R theory~\citep{anderson1982acquisition}, 
introducing four hierarchical cognitive levels with confidence-aware credit assignment for fine-grained, step-wise adaptation.

\rparagraph{RL for Language Models}
Reinforcement learning has proven effective for aligning large language models (LLMs)~\citep{ouyang2022traininglanguagemodelsfollow,rafailov2024directpreferenceoptimizationlanguage} and enhancing their reasoning capabilities~\citep{hu2025openreasonerzeroopensourceapproach, wang2025adaptivethinkingmodepolicy,xin2025scalingmultiturnoffpolicyrl}. 
Recently, group-based RL algorithms such as GRPO~\citep{shao2024deepseekmathpushinglimitsmathematical}, Dr.GRPO~\citep{liu2025understandingr1zeroliketrainingcritical}, and DAPO~\citep{yu2025dapoopensourcellmreinforcement} have been introduced. These methods estimate advantages from batches of samples without relying on critic models. While effective in single-turn reasoning tasks, they face limitations in multi-turn, agentic settings. Specifically, they distribute trajectory-level advantages uniformly across tokens, making it difficult to evaluate whether the applied cognitive pattern was appropriate at each step, especially under sparse terminal rewards. 
GiGPO~\citep{feng2025groupingrouppolicyoptimizationllm} addresses this by grouping steps via anchor states. However, state similarity alone does not guarantee cognitive relevance or appropriateness of reasoning. 
To tackle this challenge, we propose \copo, which estimates the cognitive depth at each step using action prediction confidence. This enables more precise, step-level credit assignment.

\section{Conclusion}
In this study, we addressed the inefficiency of fixed cognitive patterns in long-horizon agentic tasks. 
We introduced \framework, a framework grounded in ACT-R theory that enables agents to dynamically allocate cognitive depth across four levels, from instinctive responses to strategic planning. 
Through \cosft and \copo, we demonstrated that agents can learn to match their reasoning depth to step complexity.
Our \copo algorithm effectively solves step-level credit assignment by leveraging action prediction confidence, preventing the mode collapse observed in standard trajectory-level RL methods.
Experiments on ALFWorld and ScienceWorld confirm that \framework achieves state-of-the-art performance while reducing token consumption compared to uniform reasoning models. 
This work highlights the critical role of adaptive cognitive allocation in building efficient and effective language agents.
Future research directions include exploring adaptive mechanisms for broader ranges of tasks and investigating self-evolving cognitive hierarchies.

\bibliographystyle{plainnat}
\bibliography{custom}

\clearpage
\appendix
\label{sec:appendix}
\onecolumn
\section{Notations}
\label{app:notations}

Table~\ref{tab:notations} summarizes the core notations used in this paper.

\begin{table}[h]
    \centering
    \caption{Summary of Notations}
    \label{tab:notations}
    \vspace{2mm}
    \resizebox{0.98\linewidth}{!}{
    \begin{tabular}{p{0.15\linewidth} p{0.8\linewidth}}
        \toprule
        \multirow{2}{*}{\textbf{Symbol}} & \multicolumn{1}{c}{\textbf{Task Formulation}} \\
        \cmidrule{2-2}
        & \multicolumn{1}{c}{\textbf{Meaning}} \\
        \midrule
        $\mathcal{M}$ & Partially Observable Markov Decision Process (POMDP) tuple \\
        $\tau$ & Interaction trajectory containing instructions, observations, and actions \\
        $\pi_\theta$ & The language agent policy parameterized by $\theta$ \\
        $\mathcal{R}$ & Reward function evaluating task success \\
        \midrule
        \midrule
        \multirow{2}{*}{\textbf{Symbol}} & \multicolumn{1}{c}{\textbf{Cognitive Modeling}} \\
        \cmidrule{2-2}
        & \multicolumn{1}{c}{\textbf{Meaning}} \\
        \midrule
        $\mathcal{L}$ & Set of four cognitive levels $\{\mathcal{L}_1, \mathcal{L}_2, \mathcal{L}_3, \mathcal{L}_4\}$ derived from ACT-R theory \\
        $l_t$ & Selected cognitive level at time step $t$, where $l_t \in \mathcal{L}$ \\
        $th_t$ & Internal thinking process generated corresponding to $l_t$ \\
        $y_t$ & Structured output at step $t$, defined as $y_t = [l_t, th_t, a_t]$ \\
        $\mathcal{D}_{\rm{cog}}$ & Cognition-aware dataset used for the SFT stage \\
        \midrule
        \midrule
        \multirow{2}{*}{\textbf{Symbol}} & \multicolumn{1}{c}{\textbf{CoPO Optimization}} \\
        \cmidrule{2-2}
        & \multicolumn{1}{c}{\textbf{Meaning}} \\
        \midrule
        $A^{(i)}$ & Trajectory-level advantage calculated via standard GRPO \\
        $e_t^{(i)}$ & Cognitive group consisting of counterfactual reasoning paths at step $t$ \\
        $C_t^{(i,k)}$ & Action prediction confidence under cognitive level $k$ \\
        $g(\cdot)$ & Temperature-scaled softmax function for weight normalization \\
        $A_t^{(i, k)}$ & Confidence-aware step-level advantage for fine-grained credit assignment \\
        \bottomrule
    \end{tabular}
    }
\end{table}

\section{Task Details}
\label{app:eval_settings}

We evaluate our framework on two challenging agentic environments: \textbf{ALFWorld} and \textbf{ScienceWorld}. These benchmarks assess the agent's ability to perform long-horizon planning, reasoning, and environment interaction.

\paragraph{ALFWorld}
ALFWorld~\citep{shridhar2020alfworld} is a household environment built on TextWorld, where agents must explore rooms and apply commonsense reasoning to complete tasks. 
The action space includes operations such as picking up and placing items, observing the environment, and interacting with furniture. The environment provides feedback based on a set of predefined logical rules.
ALFWorld defines six types of tasks: Pick \& Place, Examine in Light, Clean \& Place, Heat \& Place, Cool \& Place, and Pick Two \& Place. We use the success rate as the evaluation metric. The maximum number of interaction steps is set to 30 for evaluation.

\paragraph{ScienceWorld}
ScienceWorld~\citep{wang2022scienceworlda} is a benchmark environment designed to evaluate agents' scientific reasoning abilities, based on a standard elementary science curriculum. It comprises 30 task types, including using measurement instruments and conducting mechanics experiments. The action space is task-specific, and the simulator returns the effects of each action.
Following prior work~\citep{lin2023swiftsagegenerativeagentfast}, we categorize tasks into three groups based on the average length of oracle trajectories: \textit{Short} ($^*\textit{Len} \leq 20$, avg. 11.76 steps), \textit{Medium} ($20 < {^*\textit{Len}} \leq 50$, avg. 28.58 steps), and \textit{Long} ($^*\textit{Len} > 50$, avg. 94.30 steps). These lengths reflect the oracle's optimal paths. 
We use task score as the primary evaluation metric and report the success rate as a complementary measure. The maximum number of interaction steps is limited to 100 for evaluation.

\section{Instruction Prompt Examples}
\label{app:cogrouter_prompt}
The system prompts for two agentic environments are presented in Listing~\ref{listing:prompt}. 
The output format for our \framework framework introduced in \S\ref{sec:level_design} is presented in Listing~\ref{listing:cogrouter}.
\lstset{
    backgroundcolor=\color[RGB]{245,245,244},
    breaklines=true,
    breakindent=0pt,
    basicstyle=\ttfamily\small,
    escapechar=|,  
    emph={ScienceWorld, ALFWorld, Instruction, Thought, Action},
    emphstyle={\bfseries\color{brown}},
    moredelim=[is][\color{brown}\bfseries]{@}{@} 
}
\begin{lstlisting}[caption={Prompt details for ScienceWorld and ALFWorld.},label=listing:prompt]
@ALFWorld Instruction:@
Interact with a household to solve a task. Imagine you are an intelligent agent in a household environment and your target is to perform actions to complete the task goal. At the beginning of your interactions, you will be given the detailed description of the current environment and your goal to accomplish. For each of your turn, you will be given a list of actions which you can choose one to perform in this turn. After each turn, the environment will give you immediate feedback based on which you plan your next few steps. If the environment outputs 'Nothing happened', that means the previous action is invalid and you should try more options. 
Reminder: the action must be chosen from the given available actions. Any actions except provided available actions will be regarded as illegal. 

Your current task is: @{task_description}@

Prior to this step, you have already taken {step_count} step(s). Below are the most recent {history_length} observations and the corresponding actions you took: 
@{action_history}@
You are now at step @{current_step}@ and your current observation is:  
@{current_observation}@
Your admissible actions of the current situation are: @{admissible actions}@

Now it's your turn to generate next step response. 

@ScienceWorld Instruction:@
You are an agent for science world. Every round I will give you an observation, you have to respond an action based on the observation to finish the given task. 

Here are the actions you may take:
[
{{"action": "open OBJ", "description": "open a container"}},
{{"action": "close OBJ", "description": "close a container"}},
{{"action": "activate OBJ", "description": "activate a device"}},
{{"action": "deactivate OBJ", "description": "deactivate a device"}},
{{"action": "connect OBJ to OBJ", "description": "connect electrical components"}},
{{"action": "disconnect OBJ", "description": "disconnect electrical components"}},
{{"action": "use OBJ [on OBJ]", "description": "use a device/item"}},
{{"action": "look around", "description": "describe the current room"}},
{{"action": "look at OBJ", "description": "describe an object in detail"}},
{{"action": "look in OBJ", "description": "describe a container's contents"}},
{{"action": "read OBJ", "description": "read a note or book"}},
{{"action": "move OBJ to OBJ", "description": "move an object to a container"}},
{{"action": "pick up OBJ", "description": "move an object to the inventory"}},
{{"action": "put down OBJ", "description": "drop an inventory item"}},
{{"action": "pour OBJ into OBJ", "description": "pour a liquid into a container"}},
{{"action": "dunk OBJ into OBJ", "description": "dunk a container into a liquid"}},
{{"action": "mix OBJ", "description": "chemically mix a container"}},
{{"action": "go to LOC", "description": "move to a new location"}},
{{"action": "eat OBJ", "description": "eat a food"}},
{{"action": "flush OBJ", "description": "flush a toilet"}},
{{"action": "focus on OBJ", "description": "signal intent on a task object"}},
{{"action": "wait", "description": "take no action for 10 iterations"}},
{{"action": "wait1", "description": "take no action for 1 iteration"}},
{{"action": "task", "description": "describe current task"}},
{{"action": "inventory", "description": "list your inventory"}}
]

Your current task is: @{task_description}@

Prior to this step, you have already taken {step_count} step(s). Below are the most recent {history_length} observations and the corresponding actions you took: 
@{action_history}@
You are now at step @{current_step}@ and your current observation is:  
@{current_observation}@

Now it's your turn to generate next step response. 
\end{lstlisting}
\lstset{
    backgroundcolor=\color[RGB]{245,245,244},
    breaklines=true,
    breakindent=0pt,
    basicstyle=\ttfamily\small,
    escapechar=|,  
    emph={},
    emphstyle={\bfseries\color{brown}},
    moredelim=[is][\color{brown}\bfseries]{@}{@} 
}
\begin{lstlisting}[caption={The output format for CogRouter.},label=listing:cogrouter]
There are four thinking levels:
@Level 1@ - Instinctive Response: Immediate reaction based on intuition, no analysis.
@Level 2@ - Situational Awareness: Assess current state and available actions before acting.
@Level 3@ - Experience Integration: Reflect on past actions and outcomes to inform current decisions.
@Level 4@ - Strategic Planning: Assess the task goal, past lessons, and current state to analyze the future impact of each candidate action and optimize the decision.
You must first choose an appropriate level of thinking (one of the four levels) to respond based on the given scenario. The chosen level MUST be enclosed within @<level></level>@ tags.
Next, reason step-by-step using the chosen thinking level. This reasoning process MUST be enclosed within @<think></think>@ tags. 
For Level 1 (Instinctive Response), use the fixed text: "Okay, I think I have finished thinking." For Levels 2-4, provide detailed reasoning as shown in examples. Once you've finished your reasoning, you should choose an admissible action for current step and present it within @<action></action>@ tags.

@[Output Format]@
Your output must adhere to the following format:
@EXAMPLE 1:@
<level>1</level>
<think>Okay, I think I have finished thinking.</think>
<action>your_next_action</action>

@EXAMPLE 2:@
<level>2</level>
<think>
Current state: [Current state and inventory]
Available actions: [What actions are valid right now]
Reasoning: [Choose the best action and explain why]
</think>
<action>your_next_action</action>

@EXAMPLE 3:@
<level>3</level>
<think>
Goal: [What needs to be accomplished]
Current state: [Current state and inventory]
Available actions: [What actions are valid right now]
Reflection: [How effective were recent actions, what was learned]
Reasoning: [Choose the best action based on experience]
</think>
<action>your_next_action</action>

@EXAMPLE 4:@
<level>4</level>
<think>
Goal: [What needs to be accomplished]
Current state: [Current state and inventory]
Available actions: [What actions are valid right now]
Reflection: [How effective were recent actions, what was learned]
Evaluation: [Assess the potential effectiveness of each candidate action]
Reasoning: [Choose the optimal action with strategic reasoning]
</think>
<action>your_next_action</action>
\end{lstlisting}

\clearpage

\section{Implementation Details}
\label{app:implementation}

\subsection{Training Data Collection}
\label{app:data_collection}
We construct training datasets for both \cosft and \copo using the training splits of ALFWorld (2,420 simulations) and ScienceWorld (2,120 simulations).
For \cosft, we randomly sample 500 environments from each benchmark to create a dataset with uniform coverage across the four cognitive levels. For each sampled environment, we first collect expert trajectories using GPT-4o without reasoning augmentation, then prompt GPT-4o to generate structured thinking processes based on a randomly selected cognitive level. Prompt templates for different levels are shown in Listing~\ref{lst:thinking_prompts}. This procedure yields roughly 25\% of data for each level, enabling the model to learn consistent cognitive patterns across all thinking modes. An example of the final data format is shown as follow.

\lstset{
    backgroundcolor=\color[RGB]{245,245,244},
    breaklines=true,
    breakindent=0pt,
    basicstyle=\ttfamily\small,
    emph={Example, conversations, human, gpt, from, value, instruction, system, CoSFT},
    emphstyle={\bfseries\color{brown}}
}
\begin{lstlisting}[caption={Example of \cosft training data format showing the structured output with cognitive level, thinking process, and action.}, label=cogsft:example]
CoSFT Example: 
{"conversations": 
[{"from": "human", "value": "You are an agent for science world. Every round I will give you an observation, you have to respond an action based on the observation to finish the given task.
Here are the actions you may take: [valid actions omitted]
There are four thinking levels:
Level 1 - Instinctive Response: Immediate reaction based on intuition, no analysis.
Level 2 - Situational Awareness: Assess current state and available actions before acting.
Level 3 - Experience Integration: Reflect on past actions and outcomes to inform current decisions.
Level 4 - Strategic Planning: Assess the task goal, past lessons, and current state to analyze the future impact of each candidate action and optimize the decision.
At each step, you must first choose an appropriate level of thinking (one of the four levels) to respond based on the given scenario. The chosen level MUST be enclosed within <level> </level> tags.
Next, reason step-by-step using the chosen thinking level. This reasoning process MUST be enclosed within <think> </think> tags. For Level 1 (Instinctive Response), use the fixed text: "Okay, I think I have finished thinking." For Levels 2-4, provide detailed reasoning as shown in examples. Once you've finished your reasoning, you should choose an admissible action for current step and present it within <action> </action> tags.
[Output format examples omitted]
Your current task is: Your task is to boil water. For compounds without a boiling point, combusting the substance is also acceptable. First, focus on the substance. Then, take actions that will cause it to change its state of matter. Your current observation is: Your task is to boil water. For compounds without a boiling point, combusting the substance is also acceptable. First, focus on the substance. Then, take actions that will cause it to change its state of matter.This room is called the hallway. In it, you see: a picture, a substance called air, the agent. You also see: A door to the green house (that is open), A door to the living room (that is open), A door to the art studio (that is open), A door to the kitchen (that is open), A door to the bedroom (that is open), A door to the workshop (that is open). Now it's your turn to generate next step response."}, 
{"from": "gpt", "value": "<level>1</level><think>Okay, I think I have finished thinking.</think><action>open door to kitchen</action>?}
], 
"system": ""}
\end{lstlisting}

For \copo, we use the remaining environments: 1,920 from ALFWorld and 1,620 from ScienceWorld. We adopt an online learning setup where the agent collects trajectories through environment interaction. During the cognitive group expansion process (see Section~\ref{sec:copo}), we apply the same prompt templates (Listing~\ref{lst:thinking_prompts}) to generate thinking processes at all four levels while keeping the actions fixed.

\lstset{
    backgroundcolor=\color[RGB]{245,245,244},
    breaklines=true,
    breakindent=0pt,
    basicstyle=\ttfamily\small,
    escapechar=|,  
    emph={},
    emphstyle={\bfseries\color{brown}},
    moredelim=[is][\color{brown}\bfseries]{@}{@} 
}
\begin{lstlisting}[caption={Prompt templates for generating thinking processes at different cognitive levels.}, label={lst:thinking_prompts}]
@Cognitive Level 2@
Based on the following information, generate a thinking process that leads to the next step action.

Task Description: {task_description}
Observation: {current_obs}
History of Previous Actions and Observations:
{history}
Next Action: {action}

You need to generate a thinking process following the exact format:
@<think>@
Current state: [Analyze the current environment state]
Available actions: [What actions are possible right now]
Reasoning: [Choose the best action and explain why]
@</think>@


@Cognitive Level 3@
Based on the following information, generate a thinking process that leads to the next step action.

Task Description: {task_description}
Observation: {current_obs}
History of Previous Actions and Observations:
{history}
Next Action: {action}

You need to generate a thinking process following the exact format:
@<think>@
Goal: [What needs to be accomplished]
Current state: [Analyze the current environment state]
Available actions: [What actions are possible right now]
Reflection: [How effective were recent actions, what was learned]
Reasoning: [Choose the best action based on experience]
@</think>@

@Cognitive Level 4@
Based on the following information, generate a thinking process that leads to the next step action.

Task Description: {task_description}
Observation: {current_obs}
History of Previous Actions and Observations:
{history}
Next Action: {action}

You need to generate a thinking process following the exact format:
@<think>@
Goal: [What needs to be accomplished]
Current state: [Analyze the current environment state]
Available actions: [What actions are possible right now]
Reflection: [How effective were recent actions, what was learned]
Evaluation: [Assess the potential effectiveness of each candidate action]
Reasoning: [Choose the optimal action with strategic reasoning]
@</think>@
\end{lstlisting}

\subsection{Training Details}
\label{app:training_details}

Our \copo algorithm extends group-based RL frameworks by introducing \textit{cognitive group expansion}, which enables step-wise credit assignment across different cognitive levels. Unlike GRPO, which applies trajectory-level updates, \copo generates alternate thinking processes for each decision point in successful trajectories (reward $R_i > 0$), keeping the observation $o_t$ and action $a_t$ fixed. This allows the model to learn which cognitive level best suits each state-action pair, based on observed outcomes.
To reduce computational cost, we skip cognitive group expansion when all trajectories in a group share the same outcome reward (\ie all succeed or all fail). In such cases, the trajectory-level advantage is zero, so reweighting would yield no gradient signal regardless of the expansion.

Table~\ref{tab:training_time} presents the training time over 150 iterations and the final performance of each online method (\ie GRPO, GiGPO, and \copo). 
Compared to GRPO, \copo requires approximately $1.4$--$1.5\times$ training time due to the additional expansion of the cognitive group.
However, this remains acceptable given the substantial performance gains, demonstrating a favorable trade-off between computational cost and task performance.
\begin{table}[htbp]
\setlength\tabcolsep{13pt}
\centering
\caption{Training time and performance comparison of different methods on ALFWorld and ScienceWorld.}
\scalebox{0.99}[0.99]{ 
\small
\begin{tabular}{lcccc}
\toprule
\multirow{2}{*}{\textbf{Method}} & \multicolumn{2}{c}{\textbf{ALFWorld}} & \multicolumn{2}{c}{\textbf{ScienceWorld}} \\
\cmidrule(lr){2-3} \cmidrule(lr){4-5}
& Training time & Performance & Training time & Performance \\
\midrule
\rowcolor{gray!8}
\multicolumn{5}{c}{\textit{Qwen2.5-7B-Instruct}} \\
\midrule
GRPO  & 38h 25m 11s & 83.5 & 61h 2m 16s  & 53.0 \\
GiGPO & 49h 33m 16s & 88.0 & 78h 21m 20s & 47.0 \\
CoPO  & 54h 58m 27s & 92.5 & 86h 15m 50s & 72.0 \\
\midrule
\rowcolor{gray!8}
\multicolumn{5}{c}{\textit{Llama3.1-8B-Instruct}} \\
\midrule
GRPO  & 36h 44m 38s & 84.0 & 51h 16m 3s  & 50.5 \\
GiGPO & 54h 15m 59s & 89.5 & 88h 14m 38s & 64.0 \\
CoPO  & 48h 24m 30s & 91.5 & 81h 36m 45s & 70.5 \\
\bottomrule
\end{tabular}
}
\label{tab:training_time}
\end{table}

\clearpage

\section{Extended Results}
\label{app:extended_results}

\subsection{Training Dynamics of Cognitive Level Distribution} \label{app:extended_main_results}
\begin{figure*}[ht!]
    \centering
       \vspace{-0.1cm}    \includegraphics[width=0.95\linewidth]{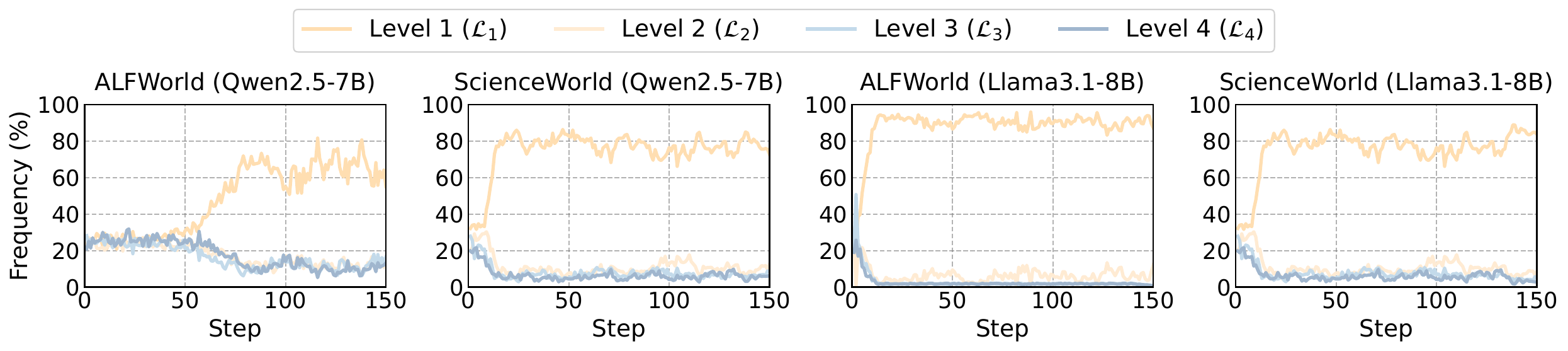}
    \caption{Dynamics of cognitive level distribution during training.}
    \label{fig:app_training_dynamic}
    \vspace{-0.36cm}
\end{figure*}

Figure~\ref{fig:app_training_dynamic} illustrates the evolution of cognitive level distribution during \copo training: the agent learns to allocate $\mathcal{L}_1$ for routine steps while preserving higher-level reasoning ($\mathcal{L}_2$-$\mathcal{L}_4$) for situations requiring deeper deliberation.

\subsection{Additional Analysis on Cognitive Level Collapse}
\label{app:cognitive_collapse}
\begin{figure}[ht!]
\centering
\includegraphics[width=0.8\linewidth]{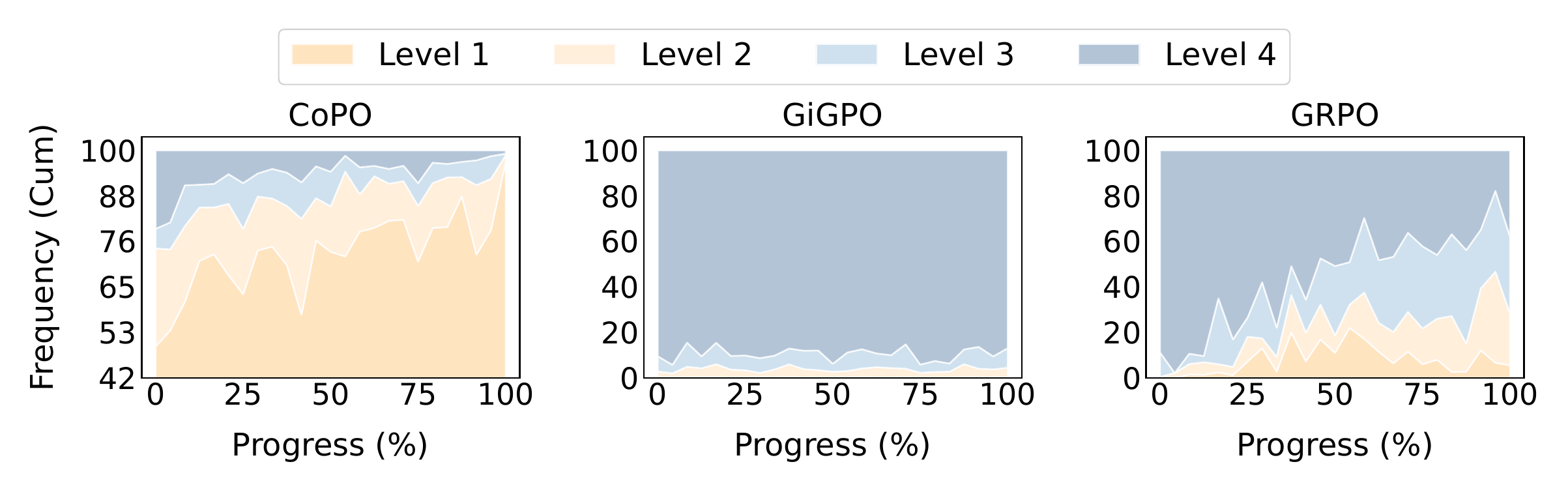}
\caption{Cognitive level distributions across trajectory progress for CoPO, GiGPO, and GRPO on ScienceWorld. Both GRPO and GiGPO collapse to predominantly $\mathcal{L}_4$ thinking throughout trajectories, while \copo exhibits stage-aware adaptation.}
\label{fig:collapse_comparison}
\end{figure}

To further understand why baseline RL methods fail to maintain adaptive cognitive allocation, we compare the cognitive level distributions across trajectory progress for \copo, GiGPO, and GRPO on ScienceWorld, using Qwen2.5-7B as the base model. 
GRPO collapses to uniform $\mathcal{L}_4$ thinking due to coarse trajectory-level credit assignment. 
Although GiGPO introduces step-level grouping, it similarly collapses, maintaining over 90\% $\mathcal{L}_4$ usage throughout the trajectory with minimal stage-wise variation (Figure~\ref{fig:collapse_comparison}). 
This collapse arises from biased credit assignment: in GiGPO's state-based grouping, steps with deeper thinking often appear in successful trajectories and receive higher advantages, regardless of whether such depth was actually necessary. 
As a result, GiGPO defaults to deeper thinking as a universally safe choice, rather than allocating cognitive effort adaptively.
In contrast, \copo employs confidence-aware reweighting to assess the appropriateness of each cognitive level at every step. 
It estimates how confidently the model predicts actions under different thinking modes, providing a fine-grained signal that guides more adaptive cognitive allocation.

\section{Case Study}
\label{app:case_study}

\begin{figure*}[t] 
\centering 
\vspace{-0.1cm} \includegraphics[width=1\linewidth]{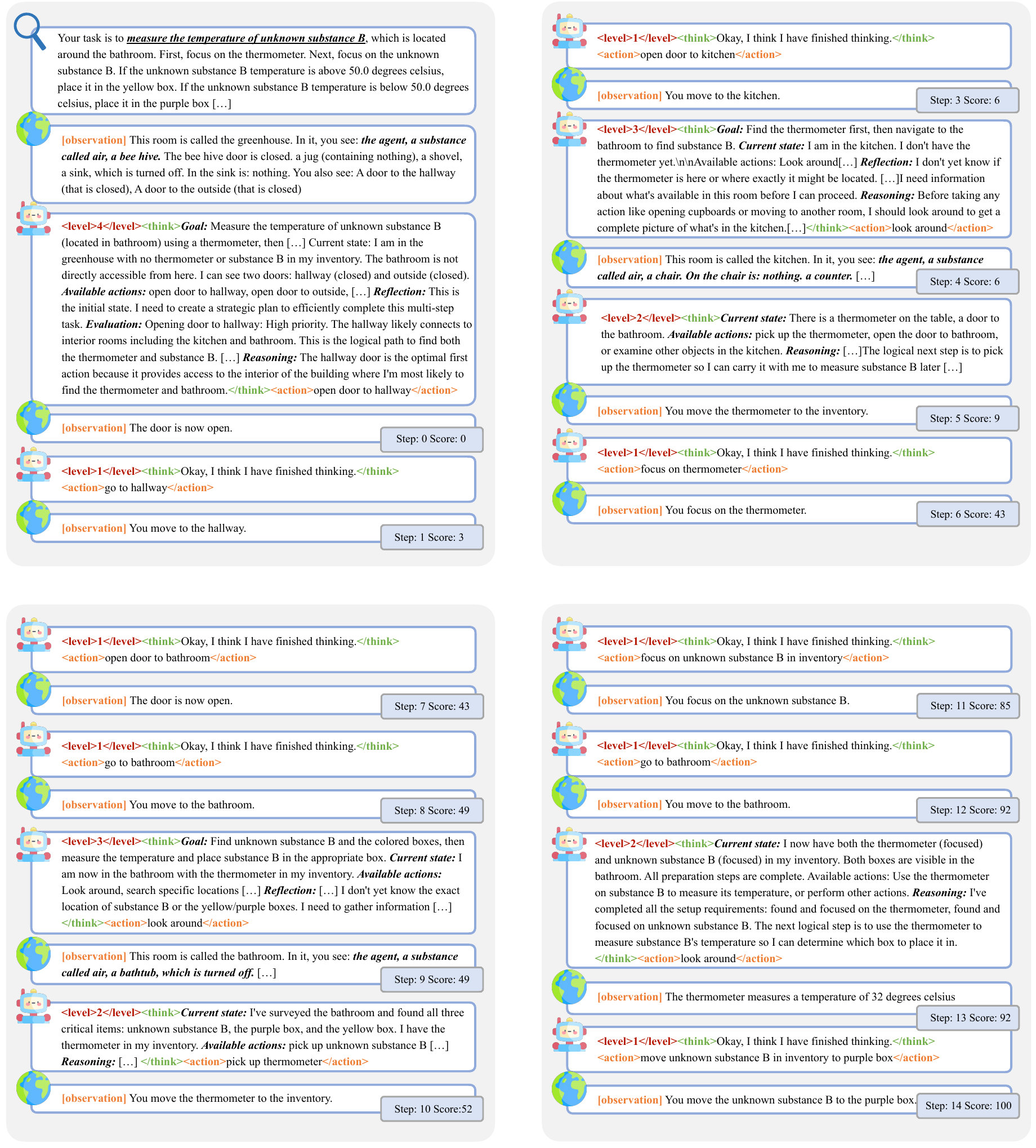} 
\caption{Case study of \framework's adaptive cognitive switching on ScienceWorld's \textit{use thermometer} task. The agent adaptively switches cognitive levels based on situational demands: employing $\mathcal{L}_4$ for initial strategic planning (step 0), $\mathcal{L}_1$ for routine navigation and manipulation (steps 6-8), $\mathcal{L}_3$ for obstacle reflection when encountering unexpected situations (step 9).}
\label{fig:case_study} 
\vspace{-0.1cm} 
\end{figure*}

We analyze a representative trajectory from ScienceWorld's \textit{use thermometer} task using Qwen2.5-7B as the base model (Figure~\ref{fig:case_study}). 
At initialization (steps 1), the agent employs $\mathcal{L}_4$ (Strategic Planning) to formulate a global plan: locate the thermometer, identify the measurement target, and execute the measurement. 
During the execution phase (steps 4-12), the agent predominantly uses $\mathcal{L}_1$ (Instinctive Response) for routine navigation and object manipulation. 
When encountering an unexpected observation at step 8 (``the door is locked''), the agent switches to $\mathcal{L}_3$ (Experience Integration) to reflect on the obstacle and adjust the plan. 
This adaptive cognitive switching demonstrates that \framework learns to allocate cognitive depth based on situational demands, achieving task success with 62\% fewer tokens than GRPO's uniform $\mathcal{L}_4$ thinking.

\clearpage

\section{Baselines}
\label{app:baseline}
\subsection{Prompt-based Methods}
\textbf{ReAct}~\citep{yao2023react} is a prompting framework that interleaves thinking processes with task-specific actions. For both ALFWorld and ScienceWorld, we prompt the model to generate a thought before each action using the format: \textit{Thought: ..., Action: ...}. This format is applied to both frontier models (GPT-4o, DeepSeek-V3, Claude-4-Sonnet, OpenAI-o3, DeepSeek-R1, Gemini-2.5-Pro) and fine-tuned open-source models (Qwen2.5-7B-Instruct, Llama3.1-8B-Instruct).
\textbf{Reflexion}~\citep{shinn2023reflexion} is a self-refinement framework that enables agents to learn from trial-and-error. After each failed episode, the agent generates a reflection summarizing the failure and proposing improvements, which is prepended to the context in subsequent trials for the same task. We allow a maximum of 3 trials per task.

\subsection{Training-based Methods}

We consider both offline and online methods using Qwen2.5-7B and Llama3.1-8B as base models. Detailed training settings are provided in Table~\ref{table:hyperparameters}.

\rparagraph{Offline Methods}  
For \textbf{SFT}, we construct a supervised dataset by randomly sampling 500 environments from the ALFWorld and ScienceWorld training sets, consistent with \cosft. For each environment, we collect expert trajectories and prompt GPT-4o to generate the corresponding reasoning steps in ReAct format, producing ReAct-style successful trajectories. These are used to fine-tune the base model via standard language modeling loss.
\textbf{ETO}~\citep{song2024trialerrorexplorationbasedtrajectory} employs Direct Preference Optimization (DPO)~\citep{rafailov2024directpreferenceoptimizationlanguage}, training on contrastive trajectory pairs. Preferred examples are taken from the SFT-generated successful trajectories, while rejected ones are sampled by rolling out the base model in the same environments.

\rparagraph{Online Methods}  
All online RL methods are trained for 150 iterations on environments not used in SFT: 1920 in ALFWorld and 1620 in ScienceWorld. In each iteration, we randomly sample 128 environments to collect trajectories for policy updates. All models are initialized from the same \cosft cold-start checkpoint for fair comparison.
\textbf{GRPO}~\citep{shao2024deepseekmathpushinglimitsmathematical} performs trajectory-level policy updates by grouping trajectories into 8 clusters.
\textbf{GiGPO}~\citep{feng2025groupingrouppolicyoptimizationllm} extends GRPO with step-level credit assignment using anchor state grouping. We use the same group size (8) for computing trajectory-level advantages. For step-level rewards, we set the weighting coefficient $\omega = 1.0$ and the discount factor $\gamma = 0.95$.
\textbf{AdaptThink}~\citep{zhang2025adaptthinkreasoningmodelslearn} teaches models to adaptively select between thinking and non-thinking modes. 
We adapt it to our multi-step agentic scenarios. 
We first perform SFT cold start by sampling 500 environments and constructing a balanced 1:1 dataset of thinking and non-thinking steps, where non-thinking uses \texttt{<think></think>} (empty think tags) to skip thinking. 
We then train with GRPO using a modified reward $r(\tau) = \frac{1}{|\tau|}\sum_{t=1}^{|\tau|} \mathbf{1}(\text{NoThink}_t) \cdot \delta + R(\tau)$, 
where $\mathbf{1}(\text{NoThink}_t)$ indicates whether step $t$ uses non-thinking mode (\ie generates \texttt{<think></think>}), and $R(\tau)$ is the task reward from~\S\ref{sec:copo}. 
The first term represents the proportion of non-thinking steps in each trajectory. We set $\delta=0.05$ and use importance sampling during training to balance exploration of both modes, with all other settings following GRPO.

\clearpage

\section{Hyperparameters}
Full hyperparameter details are provided in Table~\ref{table:hyperparameters}.
For offline methods, we use the LLaMA-Factory codebase.\footnote{\url{https://github.com/hiyouga/LLaMA-Factory}} For online methods, we implement our algorithms based on the VERL framework~\citep{Sheng_2025}.

\begin{table}[h] 
\caption{Hyperparameters for all experiments across different methods and benchmarks. All RL methods (GRPO, GiGPO, \copo) use the same hyperparameters for fair comparison.} 
\small
\centering
\setlength{\tabcolsep}{9pt}
\scalebox{0.999}{
\begin{tabular}{l|l|cc|cc} 
\toprule
& & \multicolumn{2}{c|}{Qwen2.5-7B} & \multicolumn{2}{c}{Llama3.1-8B} \\
\cline{3-6}
& & ALFWorld & ScienceWorld & ALFWorld & ScienceWorld\\
\hline
\multirow{4}{4em}{SFT} 
& Learning Rate & 2e-6 & 2e-6 & 2e-6 & 2e-6 \\
& Batch Size & 32 & 32 & 32 & 32 \\
& Number of Epoch & 3 & 3 & 3 & 3 \\
& Max Sequence Length & 8192 & 8192 & 8192 & 8192 \\
\hline
\multirow{4}{4em}{\cosft (\textsc{CoSFT}$_{\rm{exp}}$)}
& Learning Rate & 2e-6 & 2e-6 & 2e-6 & 2e-6 \\
& Batch Size & 32 & 32 & 32 & 32 \\
& Number of Epoch & 3 & 3 & 3 & 3 \\
& Max Sequence Length & 8192 & 8192 & 8192 & 8192 \\
\hline
\multirow{4}{4em}{ETO} 
& Learning Rate & 2e-6 & 2e-6 & 2e-6 & 2e-6 \\
& Batch Size & 32 & 32 & 32 & 32 \\
& Number of Epoch & 3 & 3 & 3 & 3 \\
& Max Sequence Length & 8192 & 8192 & 8192 & 8192 \\
\hline
\multirow{9}{4em}{GRPO} 
& Max Prompt Length & 18000 & 30000 & 18000 & 30000 \\
& Max Response Length & 1024 & 1024 & 1024 & 1024 \\
& Group Size & 8 & 8 & 8 & 8 \\
& Groups per Rollout & 16 & 16 & 16 & 16 \\
& KL Coefficient & 0.1 & 0.2 & 0.1 & 0.2 \\
& Learning Rate & 5e-7 & 5e-7 & 2e-7 & 2e-7 \\
& Rollout Temperature & 1.0 & 1.0 & 1.0 & 1.0 \\
& Validation Temperature & 0.4 & 0.4 & 0.4 & 0.4 \\
& Mini-batch Size & 64 & 64 & 64 & 64 \\
& Iteration & 150 & 150 & 150 & 150
\\
\hline
\multirow{9}{4em}{GiGPO} 
& Max Prompt Length & 18000 & 30000 & 18000 & 30000 \\
& Max Response Length & 1024 & 1024 & 1024 & 1024 \\
& Group Size & 8 & 8 & 8 & 8 \\
& Groups per Rollout & 16 & 16 & 16 & 16 \\
& KL Coefficient & 0.1 & 0.2 & 0.1 & 0.2 \\
& Learning Rate & 5e-7 & 5e-7 & 2e-7 & 2e-7 \\
& Rollout Temperature & 1.0 & 1.0 & 1.0 & 1.0 \\
& Validation Temperature & 0.4 & 0.4 & 0.4 & 0.4 \\
& Mini-batch Size & 64 & 64 & 64 & 64 \\
& Iteration & 150 & 150 & 150 & 150
\\
\hline
\multirow{9}{4em}{CoPO} 
& Max Prompt Length & 18000 & 30000 & 18000 & 30000 \\
& Max Response Length & 1024 & 1024 & 1024 & 1024 \\
& Group Size & 8 & 8 & 8 & 8 \\
& Groups per Rollout & 16 & 16 & 16 & 16 \\
& KL Coefficient & 0.1 & 0.2 & 0.1 & 0.2 \\
& Learning Rate & 5e-7 & 5e-7 & 2e-7 & 2e-7 \\
& Rollout Temperature & 1.0 & 1.0 & 1.0 & 1.0 \\
& Validation Temperature & 0.4 & 0.4 & 0.4 & 0.4 \\
& Mini-batch Size & 64 & 64 & 64 & 64
\\
& Iteration & 150 & 150 & 150 & 150
\\
\bottomrule
\end{tabular}
}
\label{table:hyperparameters}
\end{table}

\clearpage




\section{Necessity of Adaptive Thinking}
\label{app:ablation_adaptive}
\begin{figure}[ht!]
    \centering        \includegraphics[width=0.43\linewidth]{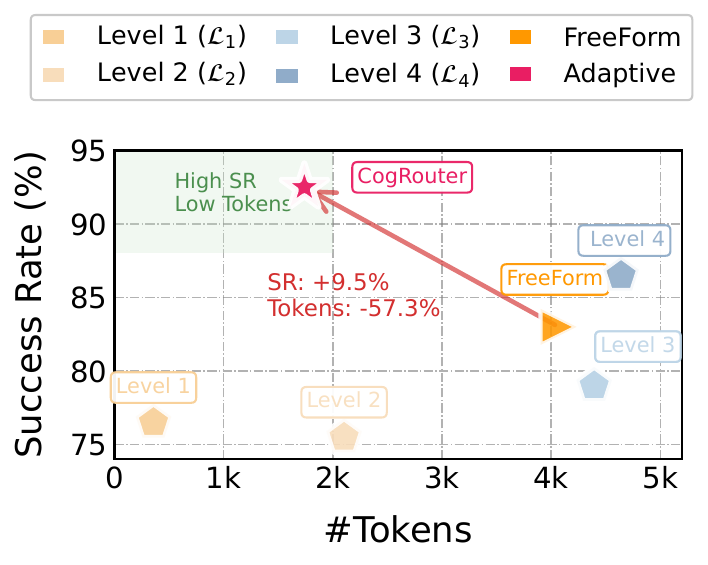}
    \caption{Comparison of cognitive configurations on ALFWorld (Qwen2.5-7B). \framework outperforms both fixed cognitive levels and unstructured reasoning, highlighting the need for adaptive thinking.}
    \label{fig:figure_05}
\end{figure}

To evaluate the importance of adaptive thinking, we compare \framework against two types of baselines on ALFWorld using Qwen2.5-7B: 
\begin{inparaenum}[\it 1)]
\item \textbf{Fixed cognitive levels}, where separate models are trained via SFT using a single reasoning format ($\mathcal{L}_1$, $\mathcal{L}_2$, $\mathcal{L}_3$, or $\mathcal{L}_4$), followed by GRPO; and 
\item \textbf{FreeForm}, trained with standard SFT using an unstructured reasoning format, also followed by GRPO.
\end{inparaenum} 
As shown in Figure~\ref{fig:figure_05}, \framework achieves a 92.5\% SR with only 1739.42 tokens, outperforming all baselines in both accuracy and efficiency. 
The fixed-level models exhibit a clear trade-off: $\mathcal{L}_1$ achieves 76.5\% SR with 357.04 tokens, while $\mathcal{L}_4$ reaches 86.5\% SR at the cost of 4640.98 tokens, still 6\% below \framework despite using 2.7$\times$ more tokens. 
FreeForm attains 81.5\% SR with 4068.88 tokens, suggesting it applies deep reasoning uniformly, without adjusting to task complexity.
These results indicate that neither shallow nor deep reasoning alone is sufficient: shallow approaches underperform on complex steps, while deep ones waste computation on simple cases. Adaptive thinking addresses this by dynamically adjusting reasoning depth to task demands.

\end{document}